\documentclass{article}

\usepackage[preprint, nonatbib]{neurips_2025}
\usepackage{comment}
\usepackage[utf8]{inputenc} 
\usepackage[T1]{fontenc}    
\usepackage{hyperref}       
\usepackage{url}            
\usepackage{booktabs}       
\usepackage{amsfonts}       
\usepackage{nicefrac}       
\usepackage{microtype}      
\usepackage{xcolor}         
\usepackage{hyperref}       
\usepackage{url}            
\usepackage{booktabs}       
\usepackage{amsfonts}       
\usepackage{nicefrac}       
\usepackage{microtype}      
\usepackage{xcolor}         
\usepackage{amsmath}
\usepackage{amssymb}
\usepackage{mathtools}
\usepackage{amsthm}
\usepackage{microtype}
\usepackage{wrapfig}
\usepackage{graphicx}
\usepackage{subfigure}
\usepackage{booktabs}
\newtheorem{theorem}{Theorem}
\newtheorem{lemma}{Lemma}

\newtheorem{remark}{Remark}

\DeclareMathOperator*{\argmax}{\arg\!\max}
\definecolor{morange}{rgb}{0.8,0.2,0}
\definecolor{mblue}{rgb}{0,0.3,1.0}
\definecolor{mred}{rgb}{0.9,0.1,0.1}
\definecolor{mpurple}{rgb}{0.5 0.1 0.7}

\title{Unlearning Algorithmic Biases over Graphs}

\author{
  O. Deniz Kose\\
  Department of Electrical Engineering and Computer Science\\
  University of California Irvine\\
  Irvine, CA, USA \\
  \texttt{okose@uci.edu} \\
  \And
   Gonzalo Mateos\\
    Department of Electrical and Computer Engineering\\
    University of Rochester\\
  Rochester, NY, USA \\
  \texttt{gmateosb@ece.rochester.edu} \\   
   \AND
   Yanning Shen\thanks{corresponding author} \\
   Department of Electrical Engineering and Computer Science\\
  University of California Irvine\\
  Irvine, CA, USA \\
  \texttt{yannings@uci.edu} \\
}

\begin{document}

\maketitle

\begin{abstract}
The growing enforcement of ``the right to be forgotten’’ regulations has propelled recent advances in certified (graph) unlearning strategies to comply with data removal requests from deployed machine learning (ML) models. Motivated by the well-documented bias amplification predicament inherent to graph data, here we take a fresh look at graph unlearning and leverage it as a bias mitigation tool. Given a pre-trained graph ML model, we develop a \emph{training-free} unlearning procedure that offers certifiable bias mitigation via a single-step Newton update on the model weights. This way, we contribute a computationally lightweight alternative to the prevalent training- and optimization-based fairness enhancement approaches, with quantifiable performance guarantees. We first develop a novel fairness-aware nodal feature unlearning strategy along with refined certified unlearning bounds for this setting, whose impact extends beyond the realm of graph unlearning. We then design structural unlearning methods endowed with principled selection mechanisms over nodes and edges informed by rigorous bias analyses.  Unlearning these judiciously selected elements can mitigate algorithmic biases with minimal impact on downstream utility (e.g., node classification accuracy). Experimental results over real networks corroborate the bias mitigation efficacy of our unlearning strategies, and delineate markedly favorable utility-complexity trade-offs relative to retraining from scratch using augmented graph data obtained via removals.
\end{abstract}

\section{Introduction}
The success of machine learning (ML) models trained on vast amounts of publicly available data is not without scrutiny due to privacy and fairness considerations. For example, UK Biobank~\cite{sudlow2015uk} data containing the genetic and medical records of approximately half a million patients \cite{ginart2019making} are often used to train classifiers in the computational biology domain. To prevent misuse of sensitive data, recent regulatory policies, such as the European Union’s General Data Protection Regulation (GDPR)~\cite{GDPR}, 
have included ``the right to be forgotten'' provisions that give subjects the right to request that their personal data be removed from datasets and ML models used by organizations~\cite{sekhari2021remember}. From an implementation perspective, 
retraining a model on the updated data after every deletion request is infeasible for most applications. This obstacle has fueled recent \emph{machine unlearning} advances, where the goal is to design efficient weight updates that result in a model without traces of data flagged for removal, and whose performance closely matches the counterfactual alternative of retraining from scratch~\cite{guo2020certified, bourtoule2021machine, izzo2021approximate, neel2021descent}. Unlearning challenges are compounded in \emph{graph} ML applications. Unlike tabular data, network datasets are characterized by two intertwined information sources: nodal features and graph connectivity. Graph components are highly correlated, and nodal representations are typically obtained via local aggregation operations~\cite{hamilton2020grlbook}, making it non-trivial to fulfill removal requests due to information ``leakage''. Recently, several graph unlearning strategies have been proposed to tackle different types of removals (e.g., feature, edge, or node unlearning)~\cite{wu2023gif, chien2022certified, wu2023certified, dong2024idea, pan2023unlearning}.

Setting aside (admittedly critical) privacy considerations and motivated by the well-documented algorithmic bias amplification predicament inherent to graph data~\cite{dai2021say, dong2023fairness1}, here we take a fresh look at \emph{graph unlearning as a bias mitigation tool}. In our treatment of (graph) ML fairness, algorithmic bias henceforth refers to undesired stereotypical correlations between model outputs and sensitive attributes, such as ethnicity, gender, or religion; see Section \ref{sec:prelim}. For computer vision and large language model (LLM) applications not related to graphs,~\cite{chen2023fast} recently showed that unlearning can be used to design efficient training-free bias mitigation strategies. The idea therein is to identify bias-inducing data samples, and unlearn them individually via a simple update to the model parameters. However, such a strategy cannot be directly used for graph data, where there are no individual data samples in transductive settings and the bias is a product of the interactions between the intertwined nodal features and graph structure~\cite{dai2021say}; see also Section \ref{sec:related} for detailed positioning relative to~\cite{chen2023fast} and other key related work. Noteworthy in- and pre-processing methods have been proposed to improve fairness in graph-based ML, which require training model parameters with fairness-related constraints~\cite{debayes}, adversarial regularization~\cite{bose2019compositional, dai2021say}, or graph editing-based data pre-processing~\cite{nifty, kose2023demystifying, spinelli2021fairdrop}). Yet, training-free post-processing algorithms have so far been underexplored in the literature~\cite{dong2023fairness1}. 

In this context, we first design principled score-based strategies to detect the feature- and topology-related elements in graph data that shape biases in a given pre-trained graph ML model. We then develop certified node feature and structural unlearning algorithms to remove such bias sources in a \emph{training-free} fashion (Section \ref{sec:method}). We thus contribute a computationally lightweight alternative to the prevalent training- and optimization-based fairness enhancement approaches, with quantifiable performance guarantees. Our contributions can be summarized as follows (see Section \ref{sec:conclusion} for limitations):\vspace{2pt}\\
\textbf{C1)} We consider a new node feature unlearning setting specific to fair graph ML, where a subset of features are removed from \emph{all} nodes. A theoretical unlearning guarantee is derived for both simple graph convolutions (SGCs) and generalized PageRank (GPR) models. Our refined bounds scale \emph{sublinearly} with the training set size, with broader impacts to unlearning of features in tabular data;\\
\textbf{C2)} We develop a training-free, efficient strategy to score candidate nodal feature removals, and theoretically establish that certified unlearning of said features can reduce algorithmic bias. Edge and node selection mechanisms are designed to unveil bias-propagating structural components, informed by rigorous bias analyses and devoid of expensive optimizations or influence function evaluations;\\
\textbf{C3)} Given a pre-trained graph ML model, we develop a \emph{training-free} unlearning procedure that offers certifiable bias mitigation via a single-step Newton update on the model weights; and\\
\textbf{C4)} Experimental results using fair graph ML benchmark networks show that our framework can effectively mitigate algorithmic bias, while enjoying markedly better performance-complexity trade-offs relative to retraining from scratch. We report fairness metric improvements of up to $~75\%$ without sacrificing utility, at a fraction of the computational complexity ($10\times$ to $20\times$ faster runtimes).

\section{Preliminaries and Problem Statement}
\label{sec:prelim}
This study focuses on machine unlearning \cite{guo2020certified}, with emphasis on \emph{fairness applications for graph data}, $\mathcal{D}_{G}:=(\mathcal{G}, \mathbf{X}, \mathbf{s}, \mathbf{y})$. The input graph is $\mathcal{G}:=(\mathcal{V}, \mathcal{E})$, where $\mathcal{V}:=$ $\left\{v_{1}, v_{2}, \ldots, v_{N}\right\}$ denotes the set of nodes and $\mathcal{E} \subseteq \mathcal{V} \times \mathcal{V}$ is the set of edges. The structural information of 
$\mathcal{G}$ is represented by the adjacency matrix  $\mathbf{A} \in  \mathbb{R}_{+}^{N \times N}$, where $A_{i j}>0$ if and only if $\left(v_{i}, v_{j}\right) \in \mathcal{E}$. Our work considers attributed graphs, where  $\mathbf{X} \in \mathbb{R}^{N \times F}$ represents the nodal features of $\mathcal{G}$. We deal with group fairness, in the context of a binary (semi-supervised) node classification task. Accordingly, \emph{sensitive attributes} are the features (such as ethnicity) with which the decisions should not be correlated for fair decision-making. Herein, the sensitive attribute is assumed to be binary and is denoted by $\mathbf{s} \in \{0,1\}^{N}$. Node $v_i$'s feature vector, sensitive attribute, and label are denoted by $\mathbf{x}_{i} \in \mathbb{R}^{F}$, $s_{i} \in \{0,1\}$, and $y_i \in \{0,1\}$, respectively. Finally, a learning algorithm over graph data is defined as a mapping $A: \mathcal{D}_{G} \mapsto \mathcal{H}_{\mathbf{w}}$ that maps a graph dataset $\mathcal{D}_{G}$ to a point $A(\mathcal{D}_{G})$ in a hypothesis space $\mathcal{H}$ parametrized by $\mathbf{w}$.

\textbf{Certified Unlearning. }  Let $\tilde{\mathcal{D}}_{G}$ denote the graph dataset obtained after removing certain nodal features or structural components from $\mathcal{D}_{G}$. Specifically, we consider feature unlearning ($\mathbf{X} \rightarrow \mathbf{\tilde{X}}$), edge unlearning ($\mathcal{E} \rightarrow \tilde{\mathcal{E}}$), and node unlearning ($\mathcal{G} \rightarrow \tilde{\mathcal{G}}$, which also implies nodal feature and incident edge removals). Given the original model $A(\mathcal{D}_G)$ parameterized by $\mathbf{w}$, unlearning algorithms aim at finding an updated model that is statistically indistinguishable from a model retrained on $\tilde{\mathcal{D}}_{G}$ from scratch. Formally, given $\epsilon, \delta>0$, an unlearning algorithm $M$ applied to $A(\mathcal{D}_{G})$ guarantees an $(\epsilon, \delta)$-certified removal for $A$, where $\mathcal{X}$ represents the space of possible datasets, if $\forall \mathcal{T} \subseteq \mathcal{H}, \mathcal{D}_{G} \subseteq \mathcal{X}$,
\begin{equation}
\label{eq:unlearning_def}
\begin{aligned}
 \mathbb{P}\big(M(A(\mathcal{D}_{G}), \mathcal{D}_{G}, \tilde{\mathcal{D}}_{G}) \in \mathcal{T}\big) \leq {}& e^\epsilon \mathbb{P}\big(A(\tilde{\mathcal{D}}_{G}) \in \mathcal{T}\big)+\delta, \\
 \mathbb{P}\big(A(\tilde{\mathcal{D}}_{G}) \in \mathcal{T}\big) \leq {}& e^\epsilon \mathbb{P}\big(M(A(\mathcal{D}_{G}), \mathcal{D}_{G},\tilde{\mathcal{D}}_{G}) \in \mathcal{T}\big)+\delta.
\end{aligned}
\end{equation}
An unlearning mechanism $M$ that satisfies $(\epsilon, \delta)$-certified removal can guarantee that the output model $M(A(\mathcal{D}_{G}), \mathcal{D}_{G}, \tilde{\mathcal{D}}_{G})$ closely approximates the model $A(\tilde{\mathcal{D}}_{G})$ obtained by retraining from scratch. These mechanisms statistically eliminate the information in the erased data $\mathcal{D}_G\setminus \tilde{\mathcal{D}}_{G}$, which we contend can be used to mitigate the algorithmic bias caused by certain ``unfair'' data components. 

\textbf{Problem Statement. } In this paper, we address the following two-pronged research question:\\
\fbox{\parbox{\textwidth}{\textit{Given a graph ML model, can we i) identify the graph data components that contribute most to said model's bias and ii) certifiably unlearn them in a training-free fashion to mitigate algorithmic bias?}}}

\section{Related Work}
\label{sec:related}
\textbf{Graph Unlearning. } Inspired by sharding-based unlearning over tabular data \cite{bourtoule2021machine}, \cite{chen2022graph, wang2023inductive, zhang2024graph} extend this approach to the graph domain, and provide efficient retraining approaches. Another group of works design training-based approaches for graph unlearning \cite{chenggnndelete, li2024towards, li2025toward, sinha2023distill, yang2023contrastive}, which, however do not provide theoretical guarantees for removal. In addition to these strategies, \cite{cong2023efficiently} projects weights into an orthogonal subspace for unlearning. Finally, influence function-based studies aim at quantifying the impact of data removal on model weights and develop efficient, training-free model updates for unlearning \cite{wu2023gif, chien2022certified, wu2023certified, dong2024idea, pan2023unlearning}. These last works provide (and inspire our own) certified removal guarantees for graph unlearning, but our distinct goal is to mitigate algorithmic bias.

\textbf{Fair Graph ML. } Designing fairness-aware learning algorithms has been attracting increasing attention \cite{dong2023fairness1, choudhary2022survey, dong2023fairness}. In terms of fairness criteria, prior work can be classified into:  i) group fairness \cite{dai2021say, rahman2019fairwalk, bose2019compositional, palowitch2020debiasing}; ii) individual fairness \cite{individual, song2022guide}; and iii) counterfactual fairness \cite{nifty, ma2022learning, guo2023towards}. Graph editing-based approaches \cite{nifty, dong2022edits, kose2022fair, kose2023demystifying, spinelli2021fairdrop} can be categorized as fair pre-processing (and in-processing) methods, while 
training-based in-processing strategies include, but are not limited to adversarial regularization \cite{dai2021say, debiasing, guo2022learning}, and Bayesian debiasing \cite{debayes}. Finally, \cite{zhang2024endowing, bose2019compositional} can be regarded as fair post-processing strategies developed for graph data, where, however, both approaches require the training of additional modules to mitigate algorithmic bias. There are also other fair post-processing methods developed specifically for link prediction \cite{masrour2020bursting} and individual fairness \cite{kang2020inform}. The setting we consider here is different from prior art, and has not been studied before. Given a pre-trained model, we develop a training-free bias mitigation algorithm that performs efficient model updates inspired by graph unlearning, under the assumption that we have access to the training data.

\textbf{Machine Unlearning and Fairness. } Although the machine unlearning problem has been studied mainly from a privacy perspective \cite{guo2020certified, li2025machine}, its impact on fairness has been getting traction recently \cite{oesterling2024fair, zhang2024forgotten, qian2024exploring}. Specifically, \cite{qian2024exploring} focuses on educational data and designs novel attacks for unlearning mechanisms that can amplify algorithmic bias. Moreover, \cite{dige2024can} leverages unlearning strategies designed to improve fairness over LLMs. Focusing on a model trained in a fairness-aware manner, \cite{oesterling2024fair} first investigates the impact of unlearning on these fair models, and then proposes a second-order (Hessian-based) update as in \cite{guo2020certified} that is compatible with a fairness-aware objective. Recent work in \cite{chen2025frog} optimizes the graph structure to alleviate the possible bias amplification caused by graph unlearning. However, the bias mitigation approach in \cite{chen2025frog} is designed as a bi-level optimization problem together with an unlearning objective, and thus markedly differs from our main goal of developing a training-free, efficient post-processing approach to improve fairness by leveraging certified graph unlearning. While~\cite{oesterling2024fair} and \cite{chen2025frog} design bias mitigation strategies to compensate for the adverse effects of unlearning \emph{randomly chosen} data samples, here we flip the script and turn this challenge into an opportunity by using graph unlearning with \emph{specialized data selection mechanisms} to mitigate bias. In terms of the adopted strategy, the closest work to ours is \cite{chen2023fast}. Specifically, \cite{chen2023fast} calculates influence functions w.r.t. counterfactual bias for training data, and then employs a second-order Newton update to unlearn data contributing most to counterfactual bias. Although this work shares a similar goal to ours --designing an efficient model update to mitigate bias-- there are several key differences: i) we leverage the distribution of data to identify biased samples, circumventing the computational burden of evaluating influence; ii) \cite{chen2023fast} relies on building counterfactual data, as it focuses on counterfactual fairness, while we are devoid of such requirements and instead consider group fairness; iii) \cite{chen2023fast} does not provide any theoretical fairness guarantee; and iv) our work \emph{distinctly focuses on graph data} with broader applicability beyond tabular data; see also Remark \ref{remark:tabular}. 

\begin{figure}[t]
    \centering
    \includegraphics[width=0.85\linewidth]{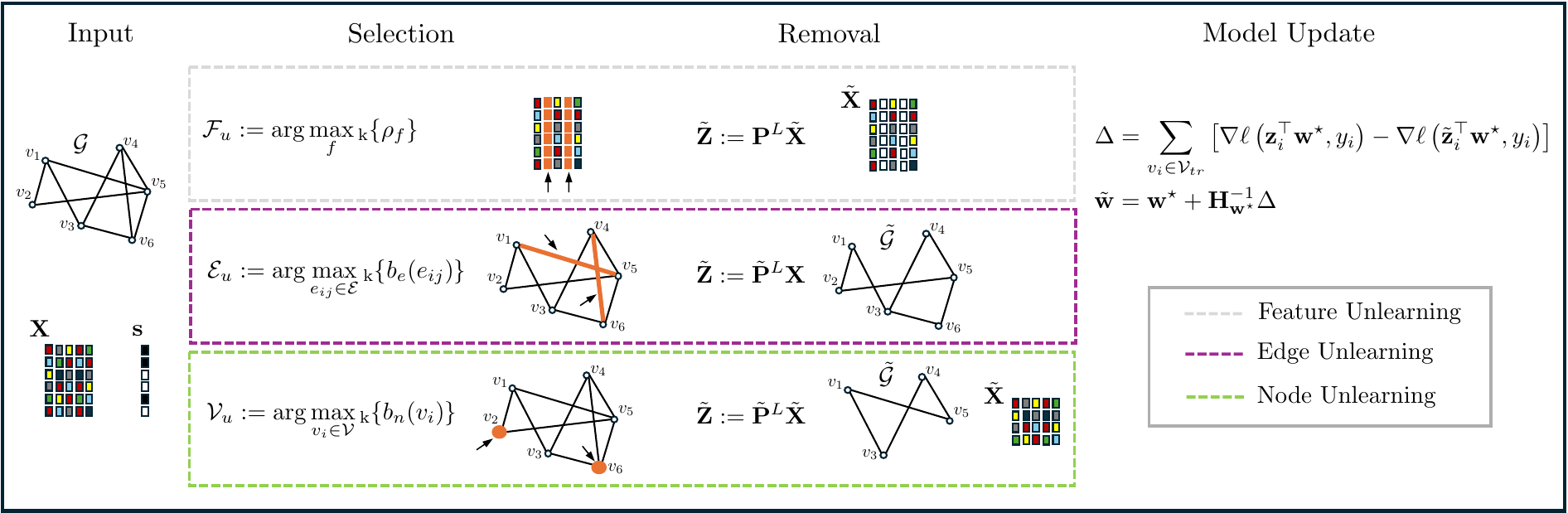} %
    \caption{Schematic of the proposed unlearning-based bias mitigation framework over graph data.}%
    \label{fig:overall}%
\end{figure}

\section{Unlearning Algorithmic Bias}
\label{sec:method}
\textbf{Learning Environment. } Building on prior work on graph unlearning for privacy \cite{chien2022certified}, we consider the SGC \cite{wu2019simplifying}, which is an architecture obtained by removing all nonlinearities from graph convolutional networks (GCNs) \cite{kipf2017semi}. The input-output relationship of SGCs can be described via $\mathbf{P}^L \mathbf{X W} := \mathbf{Z W}$, where $\mathbf{Z} \in \mathbb{R}^{N \times F}$ are the aggregated nodal representations over $L$-hop neighbors, 
by using the propagation matrix $\mathbf{P} \in\mathbb{R}^{N\times N}$,  and $\mathbf{W} \in \mathbb{R}^{F \times c}$ collects the learnable weights of the model. Although the unlearning results can be extended to multi-class classification with $c$ classes \cite{chien2022certified}, here we mainly consider a binary node classification task, thus the weights $\mathbf{W}$ can be simplified and denoted as $\mathbf{w} \in \mathbb{R}^{F \times 1}$. We adopt the left-normalized adjacency matrix with self-loops as the propagation matrix, i.e.,  $\mathbf{P} : =\bar{\mathbf{D}}^{-1} \bar{\mathbf{A}}$ with $\bar{\mathbf{A}}:=\mathbf{A}+\mathbf{I}$, where $\bar{\mathbf{D}}$ is the degree matrix for $\bar{\mathbf{A}}$.

In (semi-supervised) binary node classification the goal is to predict labels for test nodes $v_j\in\mathcal{V}_t$ given the labels $y_i\in\{0,1\}$ for training nodes $v_i\in\mathcal{V}_{tr}$, with $ |\mathcal{V}_{tr}|=m$. The model weights $\mathbf{w}$ are trained to minimize the $L_2$-regularized empirical risk, $L(\mathbf{w}; \mathcal{D}_{G})=\sum_{v_i \in \mathcal{V}_{tr} }\left(\ell\left(\mathbf{z}_i^\top\mathbf{w}, y_i\right)+\frac{\lambda}{2}\|\mathbf{w}\|^2\right)$, where $\mathbf{z}_i^\top$ denotes the $i$th row of matrix $\mathbf{Z}$ and $\lambda>0$ is a regularization parameter. For analytical purposes, $\ell\left(\mathbf{z}_{i}^{\top} \mathbf{w}, y_i\right)$ is assumed to be a convex loss function that is differentiable everywhere, which guarantees a unique minimizer for $L(\mathbf{w}; \mathcal{D}_{G})$, denoted by 
$\mathbf{w}^{\star}=A(\mathcal{D}_{G})=\operatorname{argmin}_{\mathbf{w}} L(\mathbf{w}; \mathcal{D}_{G})$. Henceforth, we let $\nabla\ell(\mathbf{z}^\top\mathbf{w},y)$ be the gradient of $\ell$ w.r.t. the weights $\mathbf{w}$, whereas $\ell^{\prime}$ and $\ell^{\prime \prime}$ stand for the first and second-order derivatives of $\ell(\cdot,\cdot)$ w.r.t. its first scalar argument.

For the graph unlearning mechanism, we employ a second-order Newton update \cite{guo2020certified}:
\begin{equation}
\label{eq:unlearning}
\begin{split}
    \Delta &= \sum_{v_i \in \mathcal{V}_{tr} } \left[\nabla \ell\left(\mathbf{z}_i^{\top} \mathbf{w}^{\star},  y_{i}\right)-\nabla \ell\left(\tilde{\mathbf{z}}_{i}^{\top}\mathbf{w}^{\star},y_{i}\right)\right],\\
    \tilde{\mathbf{w}} &= M(A(\mathcal{D}_{G}), \mathcal{D}_{G}, \tilde{\mathcal{D}}_{G}) = \mathbf{w}^{\star} + \mathbf{H}_{\mathbf{w}^{\star}}^{-1} \Delta,
\end{split}
\end{equation}
where $\mathbf{H}_{\mathbf{w}^{\star}}=\nabla^2 L(\mathbf{w}^{\star}; \tilde{\mathcal{D}}_{G})$ denotes the Hessian of $L(\cdot; \tilde{\mathcal{D}}_{G})$ at $\mathbf{w}^{\star}$, and $\tilde{\mathbf{Z}}$ represents the aggregated representations of graph data $\tilde{\mathcal{D}}_{G}$ obtained after (feature, edge, or node) removal requests. For the analytical conclusions derived and used in this study, we make the following assumptions: i) $\left\|\nabla \ell\left(\mathbf{z}_i^\top \mathbf{w}, y_i \right)\right\| \leq c$; ii) $\ell^{\prime}$ is $c_1$-bounded; iii)  $\ell^{\prime \prime}$ is $\gamma_2$-Lipschitz; and iv) $\mathbf{x}_{i} \sim \textrm{Normal}(\mathbf{0}, \boldsymbol{\Sigma})$ are i.i.d. for all $i\in\mathcal{V}$, where $\boldsymbol{\Sigma}$ is such that $\sigma^2:=\Sigma_{ii}=\Sigma_{jj}, \forall i,j$ and $\left\|\mathbf{x}_{i}\right\| \leq 1$, with high probability. For example, the logistic loss satisfies i)-iii) with $c=c_{1}=1$ and $\gamma_{2} = \frac{1}{4}$~\cite{chien2022certified}.


Considering a model trained using a noisy objective $L_{\mathbf{b}}(\mathbf{w}; \mathcal{D}_{G}) := L(\mathbf{w}; \mathcal{D}_{G})+\mathbf{b}^\top \mathbf{w}$, with $\mathbf{b}$ drawn from a Normal distribution to be fully specified later, \cite{guo2020certified} establishes that boundedness of $\|\nabla L(\tilde{\mathbf{w}}; \tilde{\mathcal{D}}_{G})\|$ guarantees $(\epsilon, \delta)$-certified removal for the Netwon update \eqref{eq:unlearning}:

\begin{theorem}[\cite{guo2020certified}]
\label{theorem:general}
Let A be the learning algorithm that returns the unique optimum of the loss $L_{\mathbf{b}}(\mathbf{w}; \mathcal{D}_{G})$, for which Assumptions i)-iii) hold. Suppose that $\|\nabla L(\tilde{\mathbf{w}}; \tilde{\mathcal{D}}_{G})\| \leq \epsilon^{\prime}$ for some computable bound $\epsilon^{\prime}>0$, independent of $\mathbf{b}$ and achieved by $M$. If $\mathbf{b} \sim \textrm{Normal}\left(\mathbf{0}, c_0 \epsilon^{\prime} / \epsilon \cdot \mathbf{I}\right)$ with $c_0>0$, then $M$ satisfies \eqref{eq:unlearning_def} with $(\epsilon, \delta)$ for algorithm A applied to $\tilde{\mathcal{D}}_{G}$, where $\delta=1.5 e^{-c_0^2 / 2}$.
\end{theorem}

Building on Theorem \ref{theorem:general}, in the sequel we develop fairness-aware unlearning requests that result in bounded $\|\nabla L(\tilde{\mathbf{w}}; \tilde{\mathcal{D}}_{G})\|$, provably. We thus contribute novel strategies guaranteeing certifiable removal of the bias propagated by unfair data points and graph (sub)structures; see also Fig. \ref{fig:overall}.

\subsection{Node Feature Unlearning}
\label{subsec:nf}
Existing literature on machine unlearning generally considers removals at the instance level, where the information requested to be erased concerns one data sample (with all its features) or a cluster of training data samples~\cite{li2025machine}. However, many studies have shown that algorithmic bias mainly stems from \emph{features} that are highly correlated with the sensitive attributes \cite{wang2022improving, kose2023demystifying}. Thus, existing instance-level unlearning strategies cannot handle more fine-grained unlearning requests at the feature level, that can be devised to improve fairness. Although feature-level unlearning is comparably understudied, there are unlearning algorithms designed for feature and label removals \cite{warnecke2021machine, xu2024don, guo2022efficient, chen2024post}. Among them, only~\cite{warnecke2021machine} provides a certified removal guarantee, in the form of a bound for $\|\nabla L(\tilde{\mathbf{w}}, \tilde{\mathcal{D}}_{G})\|$ that scales linearly with the number of data points affected by the feature removal. With a potential fairness application in mind, the features that propagate bias should be deleted from \emph{all} samples, which limits the applicability of the guarantee in \cite{warnecke2021machine} to fairly small-scale datasets. 

In this work, we consider the setting whereby $k$ nodal features are unlearned from all $N$ data points. The key issue of \emph{feature selection} for removal will be revisited and fully addressed later. Let the corresponding post-removal feature matrix be denoted by $\tilde{\mathbf{X}}$. Note that removing $k$ features from $\mathbf{X}$ leads to a reduced dimensionality model, $\mathbf{w}{'} \in \mathbb{R}^{F-k}$. Instead, we zero-out the features that will be unlearned; assumed to be the last $k$, without loss of generality. If $\mathbf{\tilde{w}}$ denotes the corresponding updated model, one has that $\tilde{\mathbf{w}}=[(\mathbf{w}^{\prime})^\top \: \mathbf{0}^\top]^\top$ for the considered linear model and the regularization term. Likewise, partition the features (columns) of $\mathbf{X} = \left[\begin{array}{ll}
\mathbf{
X}_{ns} & \mathbf{
X}_{s}
\end{array}\right] \in \mathbb{R}^{N \times F}$, and let $\tilde{\mathbf{X}} := \left[\begin{array}{ll}\mathbf{
X}_{ns} & \mathbf{0}\end{array}\right] \in \mathbb{R}^{N \times F}$, where $\mathbf{X}_{n s} \in \mathbb{R}^{N \times (F-k)} \text{are the non-sensitive ($ns$) features, and } \mathbf{X}_{s} \in \mathbb{R}^{N \times k}$.
Importantly, aggregated nodal representations $\mathbf{Z}$ also change globally post removal, becoming $\tilde{\mathbf{Z}} = \mathbf{P}^L \tilde{\mathbf{X}}=\left[\begin{array}{ll}\mathbf{
Z}_{ns} & \mathbf{0}\end{array}\right]$ for the update mechanism in SGCs. To unlearn the features, $\tilde{\mathbf{Z}}$ is fed to the model update rule \eqref{eq:unlearning}.

We establish a node feature unlearning guarantee in Theorem \ref{theorem:nodal_feat} by deriving an upper bound for $\|\nabla L(\tilde{\mathbf{w}}; \tilde{\mathcal{D}}_{G})\|$ that scales \emph{sublinearly} with the training set size $m$; see Appendix \ref{app:proof_nodal} for the proof.

\begin{theorem}
\label{theorem:nodal_feat} 
Consider nodal representations obtained via SGC. For the proposed node feature unlearning scheme under Assumptions i)-iv), 
it holds with high probability that:
\begin{equation}
\|\nabla L(\tilde{\mathbf{w}}; \tilde{\mathcal{D}}_{G})\| \leqslant \frac{\gamma_{2} }{m}\left[ \frac{2c \sqrt{F} + c_{1}\sqrt{(F-k) m}}{\lambda\sqrt{F}}\right]^{2}.
\end{equation}
\end{theorem}


As our main goal is to leverage feature unlearning to mitigate algorithmic biases, the selection of features to be unlearned is an essential component of our design. Previous works on fairness demonstrate that the correlation between features and the sensitive attribute directly influence the resulting bias \cite{zhao2022towards, kose2023demystifying}. Accordingly, we further investigate the relation between algorithmic bias and correlated features analytically. To this end, first we focus on a model trained without graph information, $\mathbf{z}_i^\top=\mathbf{x}_i^\top:=\mathbf{e}_i^\top\mathbf{X}$, where $\mathbf{e}_i$ is the $i$th canonical basic vector. We use a modified version of statistical parity (a commonly used metric for fairness evaluation \cite{sp}) that is defined based on the predicted values, $\mathbf{x}_{i}^\top\mathbf{w}^{*}$, where $\Delta^{raw}_{SP} := |\mathbb{E}_{v_{i} \sim U_{\mathcal{S}_{0}}}\left[\mathbf{x}_{i}^\top\mathbf{w}^{*} \mid v_{i} \in \mathcal{S}_0\right]-\mathbb{E}_{v_{j} \sim U_{\mathcal{S}_{1}}}\left[\mathbf{x}_{j}\top\mathbf{w}^{*} \mid v_{j} \in \mathcal{S}_1\right] |$. Here, $\mathcal{S}_{\varphi}$ denotes the set of nodes with sensitive attribute $\varphi\in\{0,1\}$, and $U_{\mathcal{S}}$ is the uniform distribution over the elements of any set $\mathcal{S}$. If $\boldsymbol{\rho}=[\rho_1,\ldots,\rho_F]^\top$ denotes the Pearson correlation \cite{lee1988thirteen} between individual input features (i.e., columns of $\mathbf{X}$) given by $\mathbf{X}\mathbf{e}_f$, $\forall f=1,\ldots,F$ and $\mathbf{s}$ ($\rho_{f}$ being the Pearson correlation between the $f$-th feature and sensitive attributes), then Theorem \ref{theorem:correlation} asserts that $\Delta^{raw}_{SP}$ can be controlled by $\boldsymbol{\rho}$. The proof of Theorem \ref{theorem:correlation} is included in Appendix \ref{app:proof_corr}.

\begin{theorem}
\label{theorem:correlation}
For binary sensitive attributes $\mathbf{s} \in \{0, 1\}^{N}$ and under Assumption iv), $\Delta^{raw}_{SP}$ is upper bounded by,
\begin{equation}
    \Delta^{raw}_{SP} \leq \frac{cN^{3/2}\bar{s} \sigma}{|\mathcal{S}_0||\mathcal{S}_1|\lambda}\|\boldsymbol{\rho}\|,
\end{equation}
where 
$\bar{s}:=\|(\mathbf{I}-\mathbf{1}\mathbf{1}^\top/N)\mathbf{s}\|$, and $\mathbf{1}=[1,\ldots,1]^\top$  denotes the all-ones vector.
\end{theorem}

Note that Theorem \ref{theorem:correlation} also applies to learning with the graph structure, where $\boldsymbol{\rho}$ becomes the vector of  correlations between the sensitive attribute and each of the rows of $\mathbf{Z} = \mathbf{P}^{L} \mathbf{X}$. Since the aggregation operation described by  $\mathbf{Z} = \mathbf{P}^{L} \mathbf{X}$ is a linear transformation of the input features, the correlation between the input features and the sensitive attributes will be equivalent to the correlation between the aggregated representations and sensitive attributes. Therefore, any (nonlinear) operation applied to $\mathbf{X}$ to reduce the correlation between features and sensitive attributes will be also effective towards diminishing the group fairness measure $\Delta^{raw}_{SP}$ defined for graph-based learning.

Theorem \ref{theorem:correlation} sheds light on the relation between algorithmic bias and nodal features that are correlated with the sensitive attributes. Removing features is in general expected to reduce $\|\boldsymbol{\rho}\|$. The extreme case would be removing all features and thus making all $\rho_f=0$, which however would also disregard all the valuable information necessary for the downstream classification task. Thus, it is imperative to design a strategy that can provide a good utility-fairness trade-off, where we can maximally remove algorithmic bias for a prescribed feature removal budget $k$. Informed by our analyses in this section, our idea is to unlearn the $k$ features in $\mathcal{F}_{u}$ that correlate most with the sensitive attributes:
\begin{equation}
\mathcal{F}_{u} :=\argmax_{f}\operatorname{_{k}}\{\rho_f\},
\end{equation}
where $\operatorname{argmax_{k}}$ selects the $k$ elements with the top-$k$ correlations. This lightweight feature selection scheme leads to lower $\|\boldsymbol{\rho}\|$ (and hence $\Delta_{SP}^{raw}$ as per Theorem \ref{theorem:correlation}) values compared to random feature selection. Our extensive empirical evaluation in Section \ref{subsec:res_feat} offers evidence supporting this claim.

\begin{remark}[Generalized PageRank-based Models (GPRs)]\label{remark:gpr}\normalfont
The certified feature unlearning guarantee in Theorem \ref{theorem:nodal_feat} also holds for GPRs, with aggregated nodal representations given by $\mathbf{Z}=\frac{1}{L+1}\left[\mathbf{X}, \mathbf{P X}, \cdots, \mathbf{P}^L \mathbf{X}\right]$.  This is stated in Theorem \ref{theorem:nodal_feat_gpr} in Appendix \ref{app:proof_nodal_gpr}, together with its the proof.
\end{remark}


\begin{remark}[Generalizability of Findings for Tabular Data]\label{remark:tabular}\normalfont
All results in this section carry over to general certified unlearning strategies for \emph{both} tabular ($\mathbf{P} = \mathbf{I}$) and graph data. To the best of our knowledge, Theorem \ref{theorem:nodal_feat} offers the first bound on $\|\nabla L(\tilde{\mathbf{w}}, \tilde{\mathcal{D}}_{G})\|$ that scales sublinearly with the number of data samples affected by unlearning. This has important and broader implications for fine-grained unlearning requests at the feature and label levels. Furthermore, the proposed feature unlearning updates can be used to mitigate algorithmic bias in a \emph{training-free} fashion, for both tabular and graph data. On top of its computational efficiency (see Section \ref{sec:exp}), a training-free post-processing strategy e.g., bypasses all known stability issues related to adversarial training for debiasing~\cite{kodali2017convergence}.
\end{remark}


\subsection{Structural Unlearning}
\label{subsec:struct}
The graph structure is a determining factor for the resulting bias in graph ML; see e.g., \cite{dai2021say, agarwal2021towards}. Such bias amplification can be attributed to the homophily principle, where nodes typically connect to other nodes that are similar to themselves. This leads to a denser connectivity between nodes with the same sensitive attribute, and hence the graph topology generally leaks sensitive information during the aggregation mechanism used to learn node representations (e.g., $\mathbf{Z}=\mathbf{P}^L\mathbf{X}$). Motivated by this, our goal in this section is to design edge and node selection mechanisms, such that forgetting said graph components can mitigate the algorithmic bias propagated by the graph connectivity.

Since we are after efficient, training-free model updates to improve fairness of pre-trained models (a post-processing operation), the efficiency of the data selection mechanisms is a key driver behind our proposed designs. Existing works that leverage unlearning for fairness \cite{chen2025frog, chen2023fast} employ selection strategies that require high computation/training cost, e.g., calculating influence functions for training data points. Although explainability methods developed for graph data \cite{ying2019gnnexplainer} can be used to find the edges and nodes that have the largest impact on a chosen bias measure, these involve computationally intensive optimizations. Instead, we deliberately focus on structural unlearning mechanisms that are training-free, efficient to implement, and grounded on rigorous motivating principles. 

Recall Theorem \ref{theorem:correlation}, which asserts that the correlation between sensitive attributes and aggregated representations can be viewed as a proxy for the resulting bias. We build on the findings in \cite{kose2023demystifying} that show $\|\boldsymbol{\rho}\|$ is a function of the graph structure, to identify bias-propagating edges and nodes in a training-free fashion. We henceforth refer to the edges that connect nodes with different sensitive attributes as \emph{inter-edges}, while \emph{intra-edges} link nodes within the same sensitive group. Specifically, for aggregated representations $\mathbf{Z} = \mathbf{P}\mathbf{X}$, \cite[Th. 1]{kose2023demystifying} establishes that $\|\boldsymbol{\rho}\| \lessapprox \max(\alpha_{1}, \alpha_{2})$, where
\begin{equation}\label{eq:tnnls}
\begin{split}
\alpha_1&:=\left|1-\frac{\left|\mathcal{S}_0^\chi\right|}{\left|\mathcal{S}_0\right|}-\frac{\left|\mathcal{S}_1^\chi\right|}{\left|S_1\right|}\right|,\\ 
\alpha_2&:=\left| 1-2 \min \left( \operatorname{mean}\left(\frac{d_m^\chi}{d_m^\chi+d_m^\omega} \right\rvert\, v_m \in \mathcal{S}_0\right), \left.\operatorname{mean}\left(\left.\frac{d_n^\chi}{d_n^\chi+d_n^\omega} \right\rvert\, v_n \in \mathcal{S}_1\right)\right) \right|.   
\end{split}
\end{equation}
Here, $\mathcal{S}_{\varphi}^\chi$ represents the set of nodes with sensitive attribute $\varphi\in\{0,1\}$ having at least one inter-edge, and $d_i^\chi$ and $d_i^\omega$ denote the numbers of inter- and intra-edges incident to $v_i$, respectively.  The characterization of the correlation in \eqref{eq:tnnls} suggests that inter- and intra-edge distributions can guide the design of training-free edge/node selection mechanisms for unlearning, the subject dealt with next. 

\textbf{Edge Unlearning. } For edge unlearning, we propose a bias score $ b_{e}: \mathcal{E} \mapsto \mathbb{R}$ to select edges in a fairness-aware unlearning strategy. Due to the homophilic (i.e., assortative) nature of connectivity in many real-world networks, the number of intra-edges is expected to be significantly larger than the number of inter-edges. Table \ref{table:stats} shows this holds true for all networks used in this study. Furthermore, as exemplified by Pokec-z \cite{dai2021say} (a dataset obtained from a real-world social network; see Section \ref{subsec:data} and Appendix \ref{app:dataset} for more details), $\left|\mathcal{S}_0^\chi\right|/\left|\mathcal{S}_0\right| = 643/6617\approx 0.10$ and $\left|\mathcal{S}_1^\chi\right|/\left|\mathcal{S}_1\right| = 603/3645\approx 0.17$. Thus, if we focus on $\alpha_{1}$ in \eqref{eq:tnnls}, reducing the number of inter-edges will make the ratios $\left|\mathcal{S}_0^\chi\right|/\left|\mathcal{S}_0\right|$ and  $\left|\mathcal{S}_1^\chi\right|/\left|\mathcal{S}_1\right|$ even smaller, resulting in higher values of $\alpha_{1}$. Regarding $\alpha_{2}$, its value can be reduced by balancing the nodal inter- and intra-degrees, and recall intra-edges are predominant. Based on these two key observations and \eqref{eq:tnnls}, our bias score \emph{prioritizes unlearning intra-edges}, where $b_{e}(e_{ij}) := \frac{\mathbb{I}\{s_{i}\equiv s_{j}\}}{\operatorname{min}(d_{i}, d_{j})}$. Here, $\mathbb{I}\{\cdot\}$ is an indicator function, which takes the value $1$ if the input condition holds, otherwise it outputs $0$; and $d_i$ is the degree of node $v_{i}$. 
The designed score incentivizes the selection of edges incident to low-degree nodes, because removing an intra-edge from a low-degree node can have a higher impact on the resulting $\frac{d_n^\chi}{d_n^\chi+d_n^\omega}$ ratio in \eqref{eq:tnnls}.
All in all, given an edge unlearning budget of $k$ edges, the proposed design unlearns the edges in set $\mathcal{E}_{u}$ given by
\begin{equation}
     \mathcal{E}_{u} :=\argmax_{e_{ij} \in \mathcal{E}}\operatorname{_{k}}\{b_e(e_{ij})\}.
\end{equation}
%
The proposed edge unlearning strategy modifies the propagation matrix $\mathbf{P}$ used for aggregation, where $\tilde{\mathbf{Z}} = \tilde{\mathbf{P}}^{L}\mathbf{X}$ is fed to the model update rule \eqref{eq:unlearning}. Our ablation studies reported in Figs. \ref{fig:edge_ablation1} and \ref{fig:edge_ablation2} further justify the design of $b_{e}$, showing the proposed bias score outperforms possible alternatives.

\textbf{Node Unlearning. } Based on the rationale used for edge unlearning, we also design a bias score to identify the nodes that influence algorithmic bias the most; hence, top candidates for removal. Let $b_{n}: \mathcal{V} \mapsto \mathbb{R}$ denote the bias score, where $b_{n}(v_{i}):= \frac{d_{i}^{\omega}}{1+d_{i}^{\chi}} \frac{1}{d_{i}}$. The term $\frac{d_{i}^{\omega}}{1+d_{i}^{\chi}}$, is directly motivated by $\alpha_{2}$ in \eqref{eq:tnnls}. Low-degree nodes are again prioritized for unlearning, as removing high-degree nodes is likely to degrade utility more severely. For an unlearning budget of $k$ nodes, we select
\begin{equation}
     \mathcal{V}_{u} := \argmax_{v_{i} \in \mathcal{V}}\operatorname{_{k}}\{b_{n}(v_{i})\}.
\end{equation}
We also empirically compared the performance of different node selection mechanisms, and our ablation results in Fig. \ref{fig:node_ablation1} corroborate the effectiveness of the proposed score $b_{n}$. Node unlearning leads to both topological- and feature-level changes, resulting in $\tilde{\mathbf{Z}} = \tilde{\mathbf{P}}^{L} \tilde{\mathbf{X}}$; see also Fig. 1.

\begin{remark}[Certified Structural Unlearning]\label{remark:certified_structural}\normalfont
The bounds for $\|\nabla L(\tilde{\mathbf{w}}; \tilde{\mathcal{D}}_{G})\|$ in \cite{chien2022certified} hold for the novel structural unlearning strategies. Thus, we remove graph components motivated by the theoretical findings in \eqref{eq:tnnls} with certifiable unlearning guarantees; see also the experiments in Sections \ref{subsec:res_edge}
-\ref{subsec:res_node}.
\end{remark}



\section{Experiments}
\label{sec:exp}

\subsection{Datasets and Experimental Setup}
\label{subsec:data}
\noindent \textbf{Datasets.} As our work approaches the graph unlearning problem from a fairness perspective, we employ commonly used real-world data for fair graph ML tasks. Specifically, the results are reported on four attributed networks: Credit Defaulter \cite{agarwal2021towards}, Recidivism \cite{agarwal2021towards},  German Credit \cite{agarwal2021towards}, Pokec-z \cite{dai2021say}. For more details on the datasets and their (e.g., inter- and intra-edge) statistics; see Appendix \ref{app:dataset}.

\noindent \textbf{Evaluation Metrics.} For node classification, we adopt accuracy as the utility measure, and we also report removal times to assess time complexity. Two quantitative measures of group fairness metrics are reported as well, namely \textit{statistical parity}: $\Delta_{S P}=|P(\hat{y}=1 \mid s=-1)-P(\hat{y}=1 \mid s=1)|$ and \textit{equal opportunity}: $\Delta_{E O}=|P(\hat{y}=1 \mid y=1, s=-1)-P(\hat{y}=1 \mid y=1, s=1)|$, where $y$ is the ground truth label, and $\hat{y}$ represents the predicted label. Lower values for $\Delta_{S P}$ and $\Delta_{E O}$ (averaged over test nodes $\mathcal{V}_t$) indicate better fairness performance \cite{dai2021say} and hence are more desirable.

\noindent \textbf{Learning Settings.} 
We test the proposed strategies using a linear GPR model with $L=3$-hop propagation. However, our unlearning guarantees extend when we use a differentially private (DP) \emph{nonlinear} DNN (e.g., a GNN) as a feature extractor, with an unlearnable linear layer~\cite{guo2020certified}. 
For more details on hyperparameters and the experimental setup, see Appendix \ref{app:hypers}.

\subsection{Node Feature Unlearning}
\label{subsec:res_feat}
We present our results for node feature unlearning and report the performance for accuracy, fairness, and runtime complexity. We employ random node feature unlearning as a baseline, where each node feature is equally likely to be chosen for removal, to demonstrate the fairness impact of our nodal feature selection strategy presented in Section \ref{subsec:nf}. Table \ref{table:comp_lp} presents comparative results for the pre-trained model and the proposed fairness-aware unlearning strategy with a budget of $k$ features. In summary, Table \ref{table:comp_lp} shows that our strategy can achieve reductions of up to $70\%$ in terms of bias metrics, without degradation in classification accuracy relative to the pre-trained model.

\begin{table}[]
	\centering
\caption{Comparative results for node feature unlearning.}
\label{table:comp_lp}
\begin{footnotesize}
\resizebox{0.95\textwidth}{!}{
\begin{tabular}{l c c c c c c }
\toprule
                                                    & \multicolumn{3}{{c}}{Credit Defaulter}& \multicolumn{3}{{c}}{German Credit}                                     \\ 
\cmidrule(r){2-4} \cmidrule(r){5-7}
                             & Acc ($\%$) & $\Delta_{S P}$ ($\%$) & $\Delta_{E O}$ ($\%$)  & Acc ($\%$) & $\Delta_{S P}$ ($\%$) & $\Delta_{E O}$ ($\%$) \\                           
 \midrule
 {Pre-trained Model} & {$67.61 \pm 0.68$} & $21.23 \pm 1.94$  & $21.72 \pm 2.86$  & {$60.10 \pm 3.18$} & $34.68 \pm 9.05$  & $33.49 \pm 12.15$ \\ 
{Random Unlearning ($k=1$)} & {$67.04 \pm 1.86$} & $21.53 \pm 3.94$  & $22.25 \pm  4.58$  & {$60.45 \pm 3.98$} & $33.53 \pm 9.97$  & $32.37 \pm  12.70$  \\ 
{Fair Unlearning ($k=1$)} & {$67.76 \pm 0.71$} & $16.06 \pm 2.01$  & $16.04 \pm  2.71$& {$60.50 \pm 3.36$} & $23.55 \pm 8.70$  & $20.54 \pm  10.44$ \\ 
{Random Unlearning ($k=5$)} & {$65.65 \pm 4.76$} & $22.31 \pm 4.67$  & $23.21 \pm  4.70$ & {$59.35 \pm 3.13$} & $32.56 \pm 10.40$  & $30.58 \pm  12.13$ \\ 
{Fair Unlearning ($k=5$)} & {$\mathbf{69.80} \pm 0.73$} & $\mathbf{10.97} \pm 1.92$  & $\mathbf{9.84} \pm  2.65$ & {$\mathbf{60.60} \pm 3.91$} & $\mathbf{9.46} \pm 6.65$  & $\mathbf{7.74} \pm  4.09$ \\ 

\midrule
       & \multicolumn{3}{{c}}{Recidivism}& \multicolumn{3}{{c}}{Pokec-z}                                    \\ 
\cmidrule(r){2-4} \cmidrule(r){5-7}
                             & Acc ($\%$) & $\Delta_{S P}$ ($\%$) & $\Delta_{E O}$ ($\%$) & Acc ($\%$) & $\Delta_{S P}$ ($\%$) & $\Delta_{E O}$ ($\%$) \\            
 \midrule
  {Pre-trained Model} & {$87.96 \pm 0.75$} & $9.64 \pm 1.17$  & $4.61 \pm 1.92$  &
  {$\mathbf{67.08} \pm 1.33$} & $5.73 \pm 1.52$  & $6.95 \pm 2.02$ \\
 
{Random Unlearning ($k=1$)} & {$88.07 \pm 1.07$} & $9.57 \pm 1.15$  & $4.85 \pm  2.16$   & {$67.08 \pm 1.31$} & $5.62 \pm 1.46$  & $6.85 \pm  2.03$  \\ 
{Fair Unlearning ($k=1$)} & {$88.14 \pm 0.70$} & $7.38 \pm 1.21$  & $\mathbf{1.97} \pm  1.61$& {$67.08 \pm 1.29$} & $4.29 \pm 1.57$  & $5.55 \pm  1.84$ \\ 

{Random Unlearning ($k=4$)} & {$79.65 \pm 14.33$} & $10.21 \pm 3.64$  & $7.14 \pm  4.82$ & {$66.92 \pm 1.06$} & $5.61 \pm 1.31$  & $6.90 \pm  2.14$ \\ 
{Fair Unlearning ($k=4$)} & {$\mathbf{89.58} \pm 0.42$} & $\mathbf{6.77} \pm 1.14$  & $2.03 \pm  1.61$ & {$66.87 \pm 1.26$} & $\mathbf{3.69} \pm 1.63$  & $\mathbf{5.38} \pm  2.16$ \\ 
\bottomrule
\end{tabular}}
\end{footnotesize}
\end{table}

Figures \ref{fig:nodal_feat1} and \ref{fig:nodal_feat2} in Appendix \ref{app:add_feat_res} depict utility and fairness performance versus the unlearning budget $k$ for the proposed feature unlearning, random feature selection, as well as retraining from scratch. The performance of the proposed update closely matches retraining, which empirically validates the unlearning guarantees in Section \ref{sec:method}. Furthermore, the proposed strategy is $17\times$, $23\times$, $17\times$, and $12\times$ faster than retraining on Credit defaulter, German Credit, Recidivism, and Pokec-z, respectively. Apparently, the proposed unlearning update incurs markedly lower computational complexity than training-based strategies. Furthermore, for Credit defaulter, unlearning bias-related features is observed to improve utility. We speculate these features might be introducing noise for the classification task specific to this dataset. Overall, the proposed strategy can enhance fairness metrics significantly (up to $~75\%$) without sacrificing utility, and at a fraction of the computational complexity. 

\begin{figure}[t]
    \centering
    \subfigure[Credit Defaulter]{{\includegraphics[width=6.85cm]{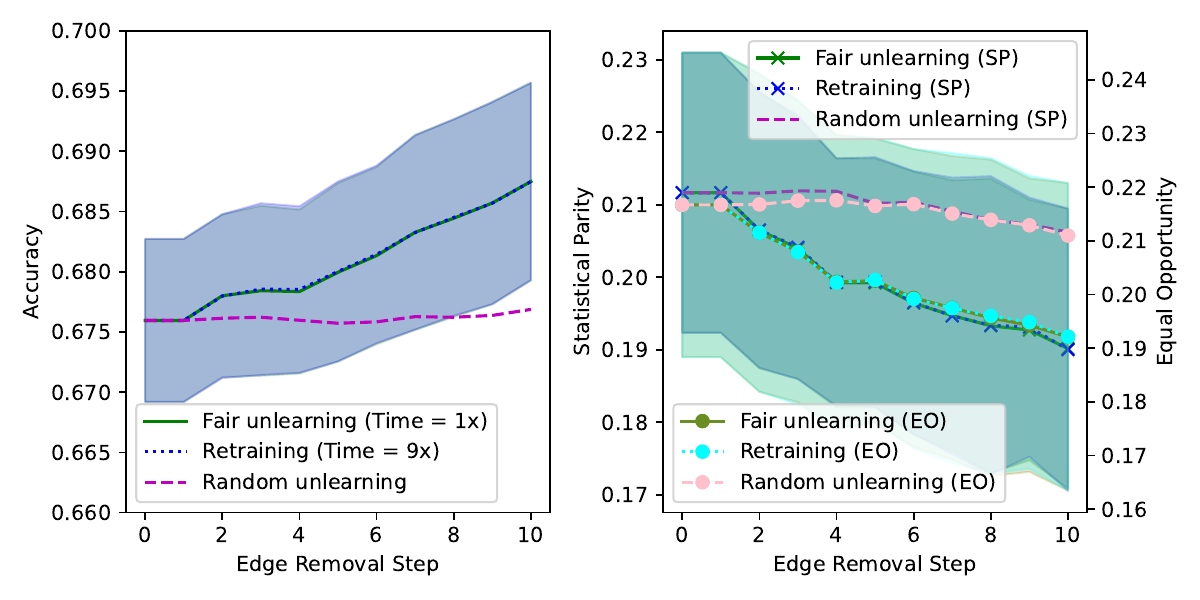} }}%
    \hspace{0.0cm}
    \subfigure[German Credit]{{\includegraphics[width=6.85cm]{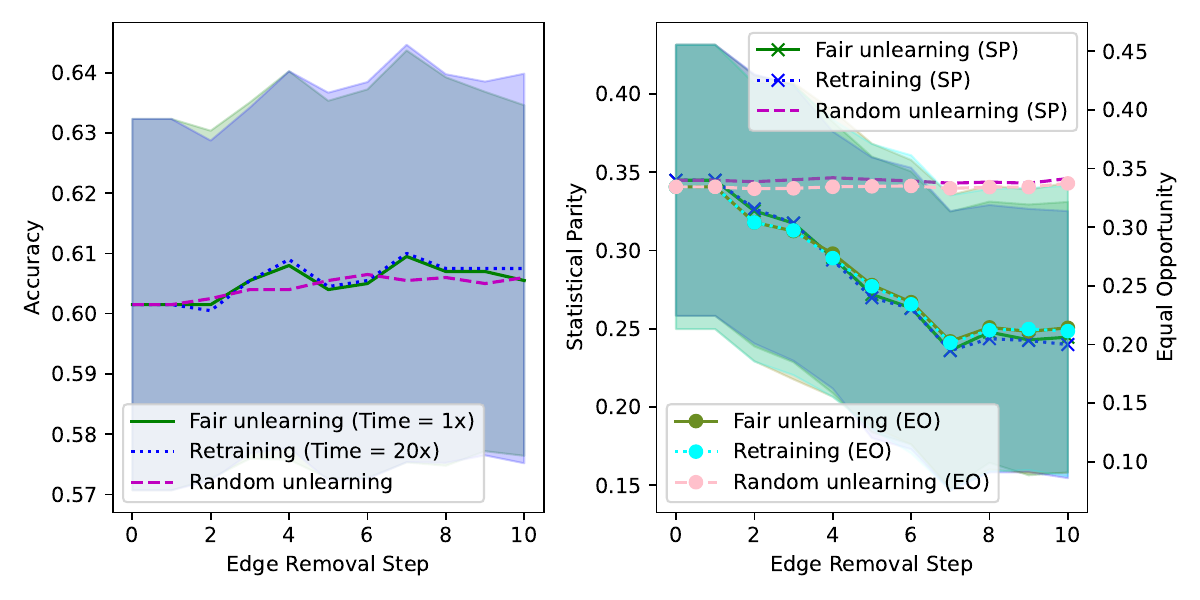} }}%
    \caption{Edge unlearning with fairness-agnostic (random) and proposed unlearning mechanisms.}%
    \label{fig:edge1}%
\end{figure}

\subsection{Edge Unlearning}
\label{subsec:res_edge}
Here we report comparative node classification results for the proposed edge unlearning scheme, employing random edge selection as the baseline. The utility, fairness, and runtime comparisons are reported in Fig. \ref{fig:edge1} for Credit Defaulter and German Credit. Note that we need to remove a significant portion of the edges to observe a nontrivial fairness impact~\cite{spinelli2021fairdrop}. For the results in Figs. \ref{fig:edge1} and \ref{fig:edge2}, our strategy unlearns $10\%$ of the edges (which can be further tuned for desirable fairness/utility trade-offs). For the Credit Defaulter network (and any other network with similar/larger sizes), unlearning $10\%$ of the edges sequentially requires $\approx 30,000$ updates, which can be time consuming. Therefore, unlearning edge mini batches might be desirable, even if certified unlearning guarantees do not directly translate into this setting. We therefore apply our unlearning algorithm using mini batches, for a total of $10$ updates, and we unlearn $1\%$ of the edges during each update.

The results in Figs. \ref{fig:edge1} and \ref{fig:edge2} exhibit behavior similar to node feature unlearning. The proposed algorithm matches the utility obtained when retraining from scratch, despite using mini matches. This matching performance is accompanied by $9\times$ and $20\times$ faster runtimes, showcasing the lightweight nature of the proposed strategy relative to training-based alternatives. Finally, the proposed edge selection mechanism can reduce the bias measures by up to $40\%$, without any noticeable loss in classification accuracy. To dissect the impact of our bias score $b_{e}$ on the fairness-utility trade-off and empirically justify our design choices, we present additional ablation studies in Appendix \ref{app:add_edge_res}.
 
\subsection{Node Unlearning} \label{subsec:res_node}
\begin{figure}[t]
    \centering
    \subfigure[Credit Defaulter]{{\includegraphics[width=6.85cm]{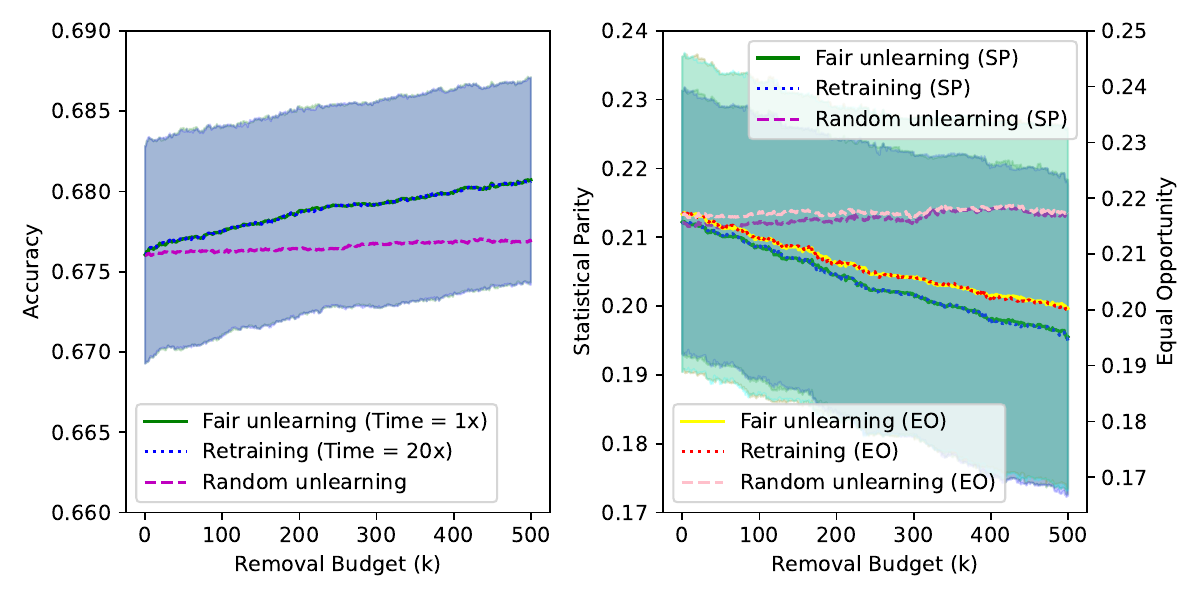} }}%
    \hspace{0.0cm}
    \subfigure[German Credit]{{\includegraphics[width=6.85cm]{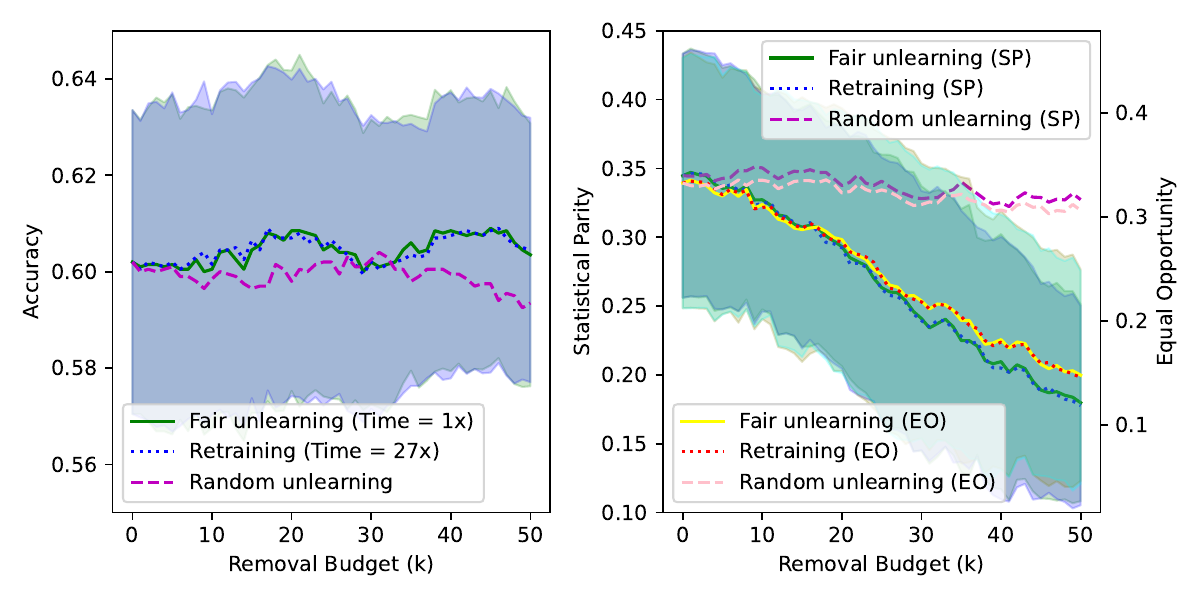} }}%
    \caption{Node unlearning with fairness-agnostic (random) and proposed unlearning mechanisms.}%
    \label{fig:node1}%
\end{figure}

For node unlearning, we employ a random node selection baseline as in Section \ref{subsec:res_feat}, where each node in $\mathcal{V}_{tr}$ is equally likely to be chosen for removal. Figure \ref{fig:node1} illustrates the performance in terms of classification accuracy, fairness, and runtime complexity. Based on the dataset sizes, we chose $k=500$ and $k=50$ for Credit Defaulter and German Credit, respectively. The results in Fig. \ref{fig:node1} again confirm the certified structural unlearning promised with improved runtime complexities, as the proposed unlearning scheme retains the performance of retraining from scratch, with a runtime at least $20\times$ faster. Once more, we observe noticeable fairness improvements and similar utility to the pre-trained model. We also conduct an ablation study, in which the fairness-utility trade-off is examined when using only the bias-related term $d_{i}^{\omega}/(1+d_{i}^{\chi})$ in $b_{n}$ as the bias score; and when node removals are based on the utility-centric score $d_{i}^{-1}$ only. Results of this ablation study are presented in Appendix \ref{app:add_node_res}, which shows that the proposed score $b_{n}$ achieves the best fairness-utility trade-off.

\section{Conclusions and Limitations}
\label{sec:conclusion}
This study takes a fresh view of graph unlearning with a focus on mitigating algorithmic bias and proposes a training-free post-processing operation to this end. We first examine a novel setting for feature unlearning that is motivated by fairness applications, where certain features are removed from \emph{all} the data points. We derive a certified unlearning guarantee that scales sublinearly with the training set size, which permeates benefits to the broader unlearning field (beyond graph data). We then propose a fairness-aware nodal feature selection mechanism for unlearning, and theoretically establish its bias mitigation merits. For structural unlearning, we rely on a formal bias analysis to guide our designs of degree-aware edge and node selection strategies. Comprehensive experiments demonstrate the effectiveness of the proposed bias mitigation approach, while also offering markedly favorable performance-complexity trade-off relative to (post removal) retraining from scratch.

\textbf{Limitations. } Our framework requires a graph ML model pre-trained in a DP-SGD manner for the theoretical unlearning guarantees to hold, limiting its applicability to a subclass of possible models. Moreover, the limitations related to Newton updates are naturally inherited here, such as the loss of certified removal guarantees for non-convex losses. We only considered unlearnable linear layers.

\bibliography{main}


\newpage
\appendix
\section{Proof of Theorem \ref{theorem:nodal_feat}}
\label{app:proof_nodal}

\textbf{Theorem. }
     For the node feature scheme proposed in Section \ref{subsec:nf}, and for $\mathbf{Z}=\mathbf{P}^L \mathbf{X}$ and $\mathbf{P}=\bar{\mathbf{D}}^{-1} \bar{\mathbf{A}}$, it holds with high probability that:
     \begin{equation}
    \left\|\nabla L\left(\tilde{\mathbf{w}}; \tilde{\mathcal{D}}_{G}\right)\right\| \leqslant \frac{\gamma_{2} }{m}\left[ \frac{2c \sqrt{F} + c_{1}\sqrt{(F-k) m}}{\lambda\sqrt{F}}\right]^{2}.
\end{equation}
     
\textbf{Goal}: We want to unlearn a subset of nodal features in this unlearning setting. Without loss of generality, we unlearn the last $k$ features in $\mathbf{X} \in \mathbb{R}^{N \times F}$, where $\tilde{\mathbf{X}}=\left[\begin{array}{ll}\mathbf{
X}_{ns} & \mathbf{0}\end{array}\right]$
$$
\mathbf{X}=\left[\begin{array}{ll}
\mathbf{
X}_{ns} & \mathbf{
X}_{s}
\end{array}\right] \text{ with } \mathbf{X}_{n s} \in \mathbb{R}^{N \times (F-k)}, \text{ and } \mathbf{X}_{s} \in \mathbb{R}^{N \times k}.
$$
For our proof, we will use Lemmas \ref{lemma:feat} and \ref{lemma:norm_z}.

\begin{lemma}
    \label{lemma:feat}
    $\left\|\nabla L\left(\tilde{\mathbf{w}}, \tilde{\mathcal{D}}_{G}\right)\right\|=\left\|\left(\mathbf{H}_{\mathbf{w}_{\eta}}-\mathbf{H}_{\mathbf{w}^*}\right) \mathbf{H}_{\mathbf{w}^{*}}^{-1} \Delta\right\|$, where $\mathbf{w}_{\eta}=\mathbf{w}^{*}+ \eta \mathbf{H}_{\mathbf{w}^{*}}^{-1} \Delta$.
\end{lemma}

\textbf{Proof of Lemma \ref{lemma:feat}}: By Taylor Theorem, there exists a $\eta \in[0,1]$, where it holds that
\begin{equation}
\begin{aligned}
\nabla L\left(\tilde{\mathbf{w}}, \tilde{\mathcal{D}}_{G}\right) & =\nabla L\left(\mathbf{w}^{*}, \tilde{\mathcal{D}}_{G}\right)+\nabla^2 L\left(\mathbf{w}^{*}+\eta \left(\tilde{\mathbf{w}}-\mathbf{w}^{*}\right), \tilde{\mathcal{D}}_{G}\right)\left(\tilde{\mathbf{w}}-\mathbf{w}^{*}\right) \\
& =\nabla L\left(\mathbf{w}^{*}, \tilde{\mathcal{D}}_{G}\right)+\mathbf{H}_{\mathbf{w}_{\eta}} \mathbf{H}_{\mathbf{w}^{*}}^{-1} \Delta .
\end{aligned}
\end{equation}
%
%
We know $\nabla L\left(\mathbf{w}^{*}, \mathcal{D}_{{G}}\right)=\sum_{v_i \in \mathcal{V}_{tr} } \nabla \ell \left(\mathbf{z}_i^{\top} \mathbf{w}^{*}, \mathbf{y}_i\right)+\lambda \mathbf{w}^* = \mathbf{0}$, as $\mathbf{w}^*$ is the unique minimizer of $\mathcal{L}\left(\mathbf{w},\mathcal{D}_{{G}}\right)$. Then, it follows that:
\begin{equation*}
\begin{aligned}
\nabla L\left(\mathbf{w}^{*}, \tilde{\mathcal{D}}_{G}\right) - \nabla L\left(\mathbf{w}^{*}, \mathcal{D}_{{G}}\right) = \nabla L\left(\mathbf{w}^{*}, \tilde{\mathcal{D}}_{G}\right) & =\sum_{v_i \in \mathcal{V}_{tr} } \nabla \ell\left(\tilde{\mathbf{z}}_i^{\top} \mathbf{w}^{*}, \mathbf{y}_i\right) -  \nabla \ell\left(\mathbf{z}_i^{\top} \mathbf{w}^{*}, \mathbf{y}_i\right) \\
& =-\Delta
\end{aligned}
\end{equation*}

We can finally conclude that $\nabla L\left(\tilde{\mathbf{w}}, \tilde{\mathcal{D}}_{G}\right) =-\Delta+\mathbf{H}_{\mathbf{w}_{\eta}} \mathbf{H}_{\mathbf{w}^{*}}^{-1} \Delta
    =\left(\mathbf{H}_{\mathbf{w}_{\eta}} - \mathbf{H}_{\mathbf{w}^{*}}\right)  \mathbf{H}_{\mathbf{w}^{*}}^{-1} \Delta.$\hfill \qed
\begin{lemma}
    \label{lemma:norm_z} 
For $\mathbf{Z} = \mathbf{P}^{L} \mathbf{X}$,  \begin{equation}
        \|\mathbf{z}_i\|_{2} \leq \max_{j} \|\mathbf{x}_j\|_{2}, \forall i.
    \end{equation}
    Thus, $\|\mathbf{z}_{i}\|_{2} \leq 1$ with high probability given that $\|\mathbf{x}_i\|_{2} \leq 1$ holds with high probability.
\end{lemma}

\textbf{Proof of Lemma \ref{lemma:norm_z}}: 
Given $\mathbf{z}_i= \sum_{v_j \in \mathcal{N}_{L}(v_{i})} \frac{[\bar{\mathbf{A}}^{L}]_{i j}}{d_{i}^{L}} \mathbf{x}_j$, we can write:
\begin{equation}
\begin{split}   
    \|\mathbf{z}_i\|_{2} &\leq \sum_{v_j \in \mathcal{N}_{L}(v_{i})}  \left| \frac{[\bar{\mathbf{A}}^{L}]_{i j}}{d_{i}^{L}} \right| \|\mathbf{x}_j\|_{2}\\
    & \leq \max_{j} \|\mathbf{x}_j\|_{2} \sum_{v_j \in \mathcal{N}_{L}(v_{i})}  \left| \frac{[\bar{\mathbf{A}}^{L}]_{i j}}{d_{i}^{L}} \right| \\
    & =  \max_{j} \|\mathbf{x}_j\|_{2} \sum_{v_j \in \mathcal{N}_{L}(v_{i})} \frac{[\bar{\mathbf{A}}^{L}]_{i j}}{d_{i}^{L}}\\
    &=  \max_{j} \|\mathbf{x}_j\|_{2} \frac{d_{i}^{L}}{d_{i}^{L}}\\
    & =  \max_{j} \|\mathbf{x}_j\|_{2}. 
\end{split}
\end{equation}
\hfil\qed
%

\textbf{Proof of Theorem \ref{theorem:nodal_feat}.} Based on Theorem \ref{theorem:general}, we aim at bounding $\left\|\nabla L\left(\tilde{\mathbf{w}}; \tilde{\mathcal{D}}_{G}\right)\right\|$. Lemma \ref{lemma:feat} proves that $ \|\nabla L\left(\tilde{\mathbf{w}}; \tilde{\mathcal{D}}_{G}\right)\|=\left\|\left(\mathbf{H}_{\mathbf{w}_{\eta}} - \mathbf{H}_{\mathbf{w}^{*}}\right)  \mathbf{H}_{\mathbf{w}^{*}}^{-1} \Delta \right\|$. By using Cauchy-Schwartz, it follows that

\begin{equation}
\label{eq:1}
\left\|\nabla L\left(\tilde{\mathbf{w}}; \tilde{\mathcal{D}}_{G}\right)\right\| \leqslant \left\|\left(\mathbf{H}_{\mathbf{w}_{\eta}} - \mathbf{H}_{\mathbf{w}^{*}}\right) \right\|\left\|\mathbf{H}_{\mathbf{w}^{*}}^{-1}\right\|\|\Delta\| .
\end{equation}

First, focus on the term $\left\|\left(\mathbf{H}_{\mathbf{w}_{\eta}} - \mathbf{H}_{\mathbf{w}^{*}}\right) \right\|$ :
\begin{equation}
\begin{aligned}
\left\|\left(\mathbf{H}_{\mathbf{w}_{\eta}} - \mathbf{H}_{\mathbf{w}^{*}}\right) \right\| & =\left\|\sum_{v_i \in \mathcal{V}_{tr} } \nabla^2 \ell\left(\tilde{\mathbf{z}}_i^{\top} \mathbf{w}_{\eta}, \mathbf{y}_i\right)-\nabla^2 \ell\left(\tilde{\mathbf{z}}_i^{\top} \mathbf{w}^{*}, \mathbf{y}_i\right)\right\| \\
& \leqslant \sum_{v_i \in \mathcal{V}_{tr} }\left\|\nabla^2 \ell\left(\tilde{\mathbf{z}}_i^{\top} \mathbf{w}_{\eta}, \mathbf{y}_i\right)-\nabla^2 \ell\left(\tilde{\mathbf{z}}_i^{\top} \mathbf{w}^{*}, \mathbf{y}_i\right)\right\| \\
& =\sum_{v_i \in \mathcal{V}_{tr} }\left\|\left[\ell^{\prime \prime}\left(\tilde{\mathbf{z}}_i^{\top} \mathbf{w}_{\eta}, \mathbf{y}_i\right)-\ell^{\prime \prime}\left(\tilde{\mathbf{z}}_i^{\top} \mathbf{w}^{*}, \mathbf{y}_i\right)\right] \tilde{\mathbf{z}}_i \tilde{\mathbf{z}}_i^{\top}\right\| \\
& \leqslant \sum_{v_i \in \mathcal{V}_{tr} } \| \ell^{\prime \prime}\left(\tilde{\mathbf{z}}_i^{\top} \mathbf{w}_{\eta}, \mathbf{y}_i\right)-\ell^{\prime \prime}\left(\tilde{\mathbf{z}}_i^{\top} \mathbf{w}^{*}, \mathbf{y}_i\right)\| \| \tilde{\mathbf{z}}_i \|^2 .
\end{aligned}
\end{equation}
Invoking Assumption iii), we have
\begin{equation}
\begin{aligned}
\left\|\left(\mathbf{H}_{\mathbf{w}_{\eta}} - \mathbf{H}_{\mathbf{w}^{*}}\right) \right\|& \leqslant \sum_{v_i \in \mathcal{V}_{tr} } \gamma_2\left\|\tilde{\mathbf{z}}_i^{\top} \mathbf{w}_{\eta}-\tilde{\mathbf{z}}_i^{\top} \mathbf{w}^{*}\right\|\left\|\tilde{\mathbf{z}}_i\right\|^2 \\
& \leqslant \sum_{v_i \in \mathcal{V}_{tr} }  \gamma_2\left\|\mathbf{w}_{\eta}-\mathbf{w}^{*}\right\|\left\|\tilde{\mathbf{z}}_i\right\|^3.
\end{aligned}
\end{equation}
Lemma \ref{lemma:norm_z} shows that $\left\|\mathbf{z}_i\right\| \leqslant 1$ is satisfied with high probability for the propagation matrix $\mathbf{P}=\bar{\mathbf{D}}^{-1} \bar{\mathbf{A}}$. Furthermore, for the proposed nodal feature unlearning, $\tilde{\mathbf{Z}}=\mathbf{P}^K \tilde{\mathbf{X}} = \mathbf{P}^{K} \left[\mathbf{X}^{n s} \quad \mathbf{0}\right]$. Thus, $z_{i j}=\tilde{z}_{i j}, \quad \forall j=1, \ldots F-k$ and $\tilde{z}_{i j}=0$ for $j=F-k+1, \ldots F$. Thus, it always holds that $\left\|\tilde{\mathbf{z}}_i\right\| \leqslant\left\|\mathbf{z}_i\right\|, \forall_i$, concluding $\left\|\tilde{\mathbf{z}}_i\right\| \leqslant 1$ with high probability. Following this, the below inequalities hold with high probabilities.
\begin{equation}
\begin{aligned}
\left\|\left(\mathbf{H}_{\mathbf{w}_{\eta}} - \mathbf{H}_{\mathbf{w}^{*}}\right) \right\| \leqslant \sum_{v_i \in \mathcal{V}_{tr} }  \gamma_2\left\|\mathbf{w}_{\eta}-\mathbf{w}^{*}\right\| & =\sum_{v_i \in \mathcal{V}_{tr} } \gamma_2\left\|\eta \mathbf{H}_{\mathbf{w}^{*}}^{-1} \Delta\right\| \\
& \leqslant \sum_{v_i \in \mathcal{V}_{tr} } \gamma_2\left\| \mathbf{H}_{\mathbf{w}^{*}}^{-1} \Delta\right\|\\
& = m \gamma_2\left\|\mathbf{H}_{\mathbf{w}^{*}}^{-1} \Delta\right\|\\
& \leqslant m \gamma_2\left\|\mathbf{H}_{\mathbf{w}^{*}}^{-1}\right\| \left\|\Delta\right\|.
\end{aligned}
\end{equation}
Finally, the following inequality is guaranteed to hold with high probability:
\begin{equation}
\label{eq:finding1}
    \left\|\nabla L\left(\tilde{\mathbf{w}}; \tilde{\mathcal{D}}_{G}\right)\right\| \leqslant \left\|\left(\mathbf{H}_{\mathbf{w}_{\eta}} - \mathbf{H}_{\mathbf{w}^{*}}\right) \right\|\left\|\mathbf{H}_{\mathbf{w}^{*}}^{-1}\right\|\|\Delta\| \leqslant   m \gamma_2 \left\|\mathbf{H}_{\mathbf{w}^{*}}^{-1}\right\|^{2} \left\|\Delta\right\|^{2}.
\end{equation}
Next, we bound $\left\|\mathbf{H}_{\mathbf{w}^{\star}}^{-1} \Delta\right\|$. Since $L\left(\cdot, \mathcal{D}^{\prime}\right)$ is $\lambda m$-strongly convex, we have $\left\|\mathbf{H}_{\mathbf{w}^{\star}}^{-1}\right\| \leq \frac{1}{\lambda m}$.

The last term we need to consider is $\|\Delta\|$ and we find
\begin{equation}
\begin{aligned}
\|\Delta\| & = \left\|\sum_{v_i \in \mathcal{V}_{tr} } \left[\nabla \ell\left(\mathbf{z}_i \mathbf{w}^{\star},  \mathbf{y}_{i}\right)-\nabla \ell\left(\tilde{\mathbf{z}}_{i}\mathbf{w}^{\star},\mathbf{y}_{i}\right)\right] \right\|\\
& =\left\|\sum_{v_i \in \mathcal{V}_{tr} } \ell^{\prime}\left(\mathbf{z}_i^{\top} \mathbf{w}^*, \mathbf{y}_i\right) \mathbf{z}_i-\ell^{\prime}\left(\tilde{\mathbf{z}}_i^{\top} \mathbf{w}^*, \mathbf{y}_i\right) \tilde{\mathbf{z}}_i\right\| \\
& \leqslant\left\|\sum_{v_i \in \mathcal{V}_{tr} }\left[\ell^{\prime}\left(\mathbf{z}_i^{\top} \mathbf{w}^*, \mathbf{y}_i\right)-\ell^{\prime}\left(\tilde{\mathbf{z}}_i^{\top} \mathbf{w}^{*}, y_i\right)\right] (\mathbf{z}_i)_{n s}\right\|+\left\|\sum_{v_i \in \mathcal{V}_{tr} } \ell^{\prime}\left(\mathbf{z}_i^{\top} \mathbf{w}^*, \mathbf{y}_i\right) (\mathbf{z}_i)_{s}\right\| \\
& \leqslant\left\|\sum_{v_i \in \mathcal{V}_{tr} } \ell^{\prime}\left(\mathbf{z}_i^{\top} \mathbf{w}^*, \mathbf{y}_i\right)(\mathbf{z}_i)_{n s} \right\| + \left\|\sum_{v_i \in \mathcal{V}_{tr} } \ell^{\prime}\left(\tilde{\mathbf{z}}_i^{\top} \mathbf{w}^{*}, y_i\right) (\mathbf{z}_i)_{n s}\right\|+\left\|\sum_{v_i \in \mathcal{V}_{tr} } \ell^{\prime}\left(\mathbf{z}_i^{\top} \mathbf{w}^*, \mathbf{y}_i\right) (\mathbf{z}_i)_{s}\right\|.
\end{aligned}
\end{equation}
First, focus on $\left\|\sum_{v_i \in \mathcal{V}_{tr} }  \ell^{\prime}\left(\mathbf{z}_i^{\top} \mathbf{w}^*, \mathbf{y}_i\right) (\mathbf{z}_i)_{ns}\right\|$. We know that $\sum_{v_i \in \mathcal{V}_{tr} } \ell^{\prime}\left(\mathbf{z}_i^{\top} \mathbf{w}^*, \mathbf{y}_i \right) z_i=-\lambda \mathbf{w}^*$, as $\mathbf{w}^{*}$ are the optimal weights. Thus,
\begin{equation}
\begin{aligned}
\left\|\sum_{v_i \in \mathcal{V}_{tr} }  \ell^{\prime}\left(\mathbf{z}_i^{\top} \mathbf{w}^*, \mathbf{y}_i\right) (\mathbf{z}_i)_{ns}\right\| &\leq \left\|\sum_{v_i \in \mathcal{V}_{tr} }  \ell^{\prime}\left(\mathbf{z}_i^{\top} \mathbf{w}^*, \mathbf{y}_i\right) \mathbf{z}_i\right\| \\
& =\lambda\left\|\mathbf{w}^*\right\| .
\end{aligned}
\end{equation}
Furthermore, by using Assumption i), it can be written that
\begin{equation}
\left\|\mathbf{w}^*\right\|=\frac{\left\|\sum_{v_i \in \mathcal{V}_{tr} } \nabla \ell\left(\mathbf{z}_i^{\top} \mathbf{w}^{*}, \mathbf{y}_i\right)\right\|}{\lambda \cdot m} \leqslant \frac{c}{\lambda}.
\end{equation}
$$
\left\|\sum_{v_i \in \mathcal{V}_{tr} }  \ell^{\prime}\left(\mathbf{z}_i^{\top} \mathbf{w}^*, \mathbf{y}_i\right) (\mathbf{z}_i)_{ns}\right\|\leqslant c .
$$
Similarly, we conclude that $\left\|\sum_{v_i \in \mathcal{V}_{tr} }  \ell^{\prime}\left(\mathbf{z}_i^{\top} \mathbf{w}^*, \mathbf{y}_i\right) (\mathbf{z}_i)_{s}\right\| \leqslant c$, leading to the upper bound on $\|\Delta\|$:
\begin{equation}
\label{eq:finding2}
\begin{aligned}
\|\Delta\|   \leqslant 2 c+ \left\|\sum_{v_i \in \mathcal{V}_{tr} } \ell^{\prime}\left(\tilde{\mathbf{z}}_i^{\top} \mathbf{w}^{*}, \mathbf{y}_i\right) (\mathbf{z}_i)_{n s}\right\|.
\end{aligned}
\end{equation}
Finally, we will focus on the term $\left\|\sum_{v_i \in \mathcal{V}_{tr} } \ell^{\prime}\left(\tilde{\mathbf{z}}_i^{\top} \mathbf{w}^{*}, \mathbf{y}_i\right) (\mathbf{z}_i)_{n s}\right\|$. As we use the SGC model \cite{wu2019simplifying} with a propagation matrix $\quad \mathbf{P}= \bar{\mathbf{D}}^{-1} \bar{\mathbf{A}}$, and $\mathbf{Z} = \mathbf{P}^{L} \mathbf{X}$,   $(\mathbf{z}_i)_{n s}= \sum_{v_j \in \mathcal{N}_{L}(v_{i})} \frac{[\bar{\mathbf{A}}^{L}]_{i j}}{d_{i}^{L}} (\mathbf{x}_j)_{n s}$, where $\mathcal{N}_{L}(v_{i})$ denotes the set containing the $L$-hop neighbors of node $v_i$, and $\operatorname{d}_i^L=\sum_{v_j \in \mathcal{N}_{L}\left(v_i\right)} [\bar{\mathbf{A}}^{L}]_{i j}$.
\begin{equation}
\begin{aligned}
\left\|\sum_{v_i \in \mathcal{V}_{tr} } \ell^{\prime}\left(\tilde{\mathbf{z}}_i^{\top} \mathbf{w}^{*}, \mathbf{y}_i\right) (\mathbf{z}_i)_{n s}\right\| &= \left\|\sum_{v_i \in \mathcal{V}_{tr} } \ell^{\prime}\left(\tilde{\mathbf{z}}_i^{\top} \mathbf{w}^{*}, y_i\right) (\mathbf{z}_i)_{n s}\right\|\\
&= \left\|\sum_{v_i \in \mathcal{V}_{tr} } \ell^{\prime}\left(\tilde{\mathbf{z}}_i^{\top} \mathbf{w}^{*}, \mathbf{y}_i\right) \sum_{v_j \in \mathcal{N}_{L}(v_{i})} \frac{[\bar{\mathbf{A}}^{L}]_{i j}}{d_{i}^{L}} (\mathbf{x}_j)_{n s}\right\| \\
&=  \left\|\sum_{v_i \in \mathcal{V}_{tr} } \sum_{v_j \in \mathcal{N}_{L}(v_{i})} \frac{\ell^{\prime}\left(\tilde{\mathbf{z}}_i^{\top} \mathbf{w}^{*}, \mathbf{y}_i\right) [\bar{\mathbf{A}}^{L}]_{i j}}{d_{i}^{L}} (\mathbf{x}_j)_{n s}\right\| 
\end{aligned}
\end{equation}
Based on Assumption iv), the following holds:
\begin{equation}
    \sum_{v_i \in \mathcal{V}_{tr} } \sum_{v_j \in \mathcal{N}_{L}(v_{i})} \frac{\ell^{\prime}\left(\tilde{\mathbf{z}}_i^{\top} \mathbf{w}^{*}, \mathbf{y}_i\right) [\bar{\mathbf{A}}^{L}]_{i j}}{d_{i}^{L}} (\mathbf{x}_j)_{n s} \sim  N\left(\mathbf{0},  \sum_{v_i \in \mathcal{V}_{tr} } \sum_{v_j \in \mathcal{N}_{L}(v_{i})} \left[ \frac{\ell^{\prime}\left(\tilde{\mathbf{z}}_i^{\top} \mathbf{w}^{*}, \mathbf{y}_i\right) [\bar{\mathbf{A}}^{L}]_{i j}}{d_{i}^{L}}\right]^{2} \boldsymbol{\Sigma_{ns}} \right).
\end{equation}
Let $\mathbf{o}:=   \sum_{v_i \in \mathcal{V}_{tr} } \sum_{v_j \in \mathcal{N}_{L}(v_{i})} \frac{\ell^{\prime}\left(\tilde{\mathbf{z}}_i^{\top} \mathbf{w}^{*}, \mathbf{y}_i\right) [\bar{\mathbf{A}}^{L}]_{i j}}{d_{i}^{L}} (\mathbf{x}_j)_{n s}$, where we showed $\mathbf{o} \sim  N\left(\mathbf{0},  \sum_{v_i \in \mathcal{V}_{tr} } \sum_{v_j \in \mathcal{N}_{L}(v_{i})} \left[ \frac{\ell^{\prime}\left(\tilde{\mathbf{z}}_i^{\top} \mathbf{w}^{*}, \mathbf{y}_i\right) [\bar{\mathbf{A}}^{L}]_{i j}}{d_{i}^{L}}\right]^{2} \boldsymbol{\Sigma_{ns}}\right)$. Thus, each element of $\mathbf{o}$ is a Gaussian RV, $o_{m} \sim \textrm{Normal}(\mathbf{0}, \sum_{v_i \in \mathcal{V}_{tr} } \sum_{v_j \in \mathcal{N}_{L}(v_{i})} \left[ \frac{\ell^{\prime}\left(\tilde{\mathbf{z}}_i^{\top} \mathbf{w}^{*}, \mathbf{y}_i\right) [\bar{\mathbf{A}}^{L}]_{i j}}{d_{i}^{L}}\right]^{2} \Sigma_{mm}), \forall l={1, \dots, F-k}$. Let $\sigma^{2}$ denote each diagonal entries in $\boldsymbol{\Sigma}$. Therefore, the following inequalities hold with a probability of $0.9999994$:
\begin{equation}
\label{eq:ineq}
    -5 \sqrt{\sum_{v_i \in \mathcal{V}_{tr} } \sum_{v_j \in \mathcal{N}_{L}(v_{i})} \left[ \frac{\ell^{\prime}\left(\tilde{\mathbf{z}}_i^{\top} \mathbf{w}^{*}, \mathbf{y}_i\right) [\bar{\mathbf{A}}^{L}]_{i j}}{d_{i}^{L}}\right]^{2} \sigma^{2}} \leq o_{m} \leq 5 \sqrt{\sum_{v_i \in \mathcal{V}_{tr} } \sum_{v_j \in \mathcal{N}_{L}(v_{i})} \left[ \frac{\ell^{\prime}\left(\tilde{\mathbf{z}}_i^{\top} \mathbf{w}^{*}, \mathbf{y}_i\right) [\bar{\mathbf{A}}^{L}]_{i j}}{d_{i}^{L}}\right]^{2}\sigma^{2}}.
\end{equation}
We can use the inequalities in \eqref{eq:ineq} to bound $\|\mathbf{o}\|$, where the following upper bound holds with a probability of $0.9999994^{F-k}$ ($F$ takes a maximum value of $59$ for the real networks we use, leading to an overall probability larger than $0.99$):
\begin{equation}
\label{eq:2}
    \|\mathbf{o}\| \leq \sqrt{25(F-k) \sum_{v_i \in \mathcal{V}_{tr} } \sum_{v_j \in \mathcal{N}_{L}(v_{i})} \left[ \frac{\ell^{\prime}\left(\tilde{\mathbf{z}}_i^{\top} \mathbf{w}^{*}, \mathbf{y}_i\right) [\bar{\mathbf{A}}^{L}]_{i j}}{d_{i}^{L}}\right]^{2} \sigma^{2} }.
\end{equation}
Based the definition of $\operatorname{d}_i^L=\sum_{v_j \in \mathcal{N}_{K}\left(v_i\right)} [\bar{\mathbf{A}}^{L}]_{i j}$, it follows that   $[\operatorname{d}_i^L]^{2}=\left[\sum_{v_j \in \mathcal{N}_{L}\left(v_i\right)} [\bar{\mathbf{A}}^{L}]_{i j}\right]^{2}$. As $[\bar{\mathbf{A}}^{L}]_{i j}>0$ for the adjacency matrix, it follows that $[\operatorname{d}_i^L]^{2} \geq \sum_{v_j \in \mathcal{N}_{L}\left(v_i\right)} [[\bar{\mathbf{A}}^{L}]_{i j}]^{2}$. Accordingly, \eqref{eq:2} can be rewritten as:
\begin{equation}
\begin{split}
    \|\mathbf{o}\| &\leq \sqrt{25(F-k) \sum_{v_i \in \mathcal{V}_{tr} } \sum_{v_j \in \mathcal{N}_{L}(v_{i})} \left[ \frac{\ell^{\prime}\left(\tilde{\mathbf{z}}_i^{\top} \mathbf{w}^{*}, \mathbf{y}_i\right) [\bar{\mathbf{A}}^{L}]_{i j}}{d_{i}^{L}}\right]^{2} \sigma^{2}},\\
    & = \sqrt{25(F-k) \sum_{v_i \in \mathcal{V}_{tr} } \sum_{v_j \in \mathcal{N}_{L}(v_{i})} \frac{([\bar{\mathbf{A}}^{L}]_{i j})^{2}}{(d_{i}^{L})^{2}}(\ell^{\prime}\left(\tilde{\mathbf{z}}_i^{\top} \mathbf{w}^{*}, \mathbf{y}_i\right))^{2}\sigma^{2}} \\
    & \leq \sqrt{25(F-k) \sum_{v_i \in \mathcal{V}_{tr} } \frac{(d_{i}^{L})^{2}}{(d_{i}^{L})^{2}}(\ell^{\prime}\left(\tilde{\mathbf{z}}_i^{\top} \mathbf{w}^{*}, \mathbf{y}_i\right))^{2}\sigma^{2}}\\
    & = \sqrt{25(F-k) \sum_{v_i \in \mathcal{V}_{tr} }(\ell^{\prime}\left(\tilde{\mathbf{z}}_i^{\top} \mathbf{w}^{*}, \mathbf{y}_i\right))^{2}\sigma^{2}}.
\end{split}    
\end{equation}
Furthermore, by using Assumption ii), we can conclude that:

\begin{equation}
    \label{eq:finding8}
     \|\mathbf{o}\| \leq 5 c_{1} \sqrt{(F-k) m \sigma^{2}}.
\end{equation}

Note that the assumption that $\|\mathbf{x}_{i}\| \leq 1$ holds with high probability has an implication on the $\sigma$ value. For any $x_{im}$, the following holds with a probability $0.9999994$:
\begin{equation}
    -5\sigma \leq x_{im} \leq 5\sigma.
\end{equation}
Thus, $\|\mathbf{x}_{i}\| \leq 5\sigma \sqrt{F}$ with probability $0.9999994^{F} > 0.99$ for the maximum value of $F=59$ for the real networks we use. Therefore, our assumption on the norm of $\mathbf{x}_{i}$ implies that $5\sigma \sqrt{F} \approx 1$. It follows with high probability that:
\begin{equation}
    \label{eq:finding3}
     \|\mathbf{o}\| \leq  c_{1} \sqrt{\frac{(F-k)}{F} m}.
\end{equation}
By combining findings in \eqref{eq:finding1}, \eqref{eq:finding2}, \eqref{eq:finding3}, we obtain
\begin{equation}
\begin{split}
    \left\|\nabla L\left(\tilde{\mathbf{w}}, \tilde{\mathcal{D}}_{G}\right)\right\| &\leqslant m \gamma_{2} \left[ \frac{2c\sqrt{F} + 5c_{1}\sqrt{(F-k) m}}{\lambda m}\right]^{2},\\
    &=  \frac{\gamma_{2} }{m}\left[ \frac{2c \sqrt{F} + c_{1}\sqrt{(F-k) m}}{\lambda\sqrt{F}}\right]^{2}
\end{split}
\end{equation}
and the proof is concluded.\hfill\qed

\section{Proof of Theorem \ref{theorem:correlation}}
\label{app:proof_corr}

Theorem \ref{theorem:correlation} asserts that $\Delta^{raw}_{SP} \leq \frac{cN^{3/2}\bar{s} \sigma}{|\mathcal{S}_0||\mathcal{S}_1|\lambda}\|\boldsymbol{\rho}\|$. Here, we will first focus on the $\Delta^{raw}_{SP}$ term. 
\begin{equation}
\begin{split}
    \Delta^{raw}_{SP} &= \left|\mathbb{E}_{v_{i} \sim U_{\mathcal{S}_{0}}}\left[\mathbf{x}_{i}^{\top}\mathbf{w}^{*} \mid v_{i} \in \mathcal{S}_0\right]-\mathbb{E}_{v_{j} \sim U_{\mathcal{S}_{1}}}\left[\mathbf{x}_{j}^{\top}\mathbf{w}^{*} \mid v_{j} \in \mathcal{S}_1\right] \right|,\\
    &= \left| \frac{1}{|\mathcal{S}_{0}|} \sum_{v_i \in \mathcal{S}_{0} } \mathbf{x}_{i}^{\top}\mathbf{w}^{*} - \frac{1}{|\mathcal{S}_{0}|} \sum_{v_j \in \mathcal{S}_{1} } \mathbf{x}_{j}^{\top}\mathbf{w}^{*}  \right|,\\
    &= \left| \left(\frac{1}{|\mathcal{S}_{0}|} \sum_{v_i \in \mathcal{S}_{0} } \mathbf{x}_{i} - \frac{1}{|\mathcal{S}_{0}|} \sum_{v_j \in \mathcal{S}_{1} } \mathbf{x}_{j} \right)^{\top} \mathbf{w}^{*}  \right|,\\
    & \leq \left\| \frac{1}{|\mathcal{S}_{0}|} \sum_{v_i \in \mathcal{S}_{0} } \mathbf{x}_{i} - \frac{1}{|\mathcal{S}_{0}|} \sum_{v_j \in \mathcal{S}_{1} } \mathbf{x}_{j}  \right\| \left\|\mathbf{w}^{*} \right\|.
\end{split}
\end{equation}
We have shown that $\|\mathbf{w}^{*}\| \leq \frac{c}{\lambda}$ in Appendix \ref{app:proof_nodal}. Thus, it follows that:
\begin{equation}
\label{eq:findind_2_1}
 \Delta^{raw}_{SP} \leq \frac{c}{\lambda} \left\| \frac{1}{|\mathcal{S}_{0}|} \sum_{v_i \in \mathcal{S}_{0} } \mathbf{x}_{i} - \frac{1}{|\mathcal{S}_{0}|} \sum_{v_j \in \mathcal{S}_{1} } \mathbf{x}_{j}  \right\|.
\end{equation}
By Assumption iv) we have $\mathbf{x}_{i} \sim \textrm{Normal}(\mathbf{0}, \boldsymbol{\Sigma})$ are i.i.d. for all $i\in\mathcal{V}$, where $\boldsymbol{\Sigma}$ is such that $\sigma^2:=\Sigma_{ii}=\Sigma_{jj}, \forall i, j$ and $\left\|\mathbf{x}_{i}\right\| \leq 1$, with high probability. Let $\bar{s}:=\|(\mathbf{I}-\mathbf{1}\mathbf{1}^\top/N)\mathbf{s}\|$. Then, each feature $x_{il}\sim \textrm{Normal}(0, \sigma^2)$), and the Pearson correlation between a feature $\mathbf{x}_{:l}$ and sensitive attributes $\mathbf{s}$ follows as:
\begin{equation}
\begin{split}
    \rho_{l} &= \frac{1}{ \bar{s} \sqrt{N \sigma^2}} \left(\sum_{v_i \in \mathcal{S}_{0} } (0-\bar{s})(x_{il}-\bar{x}_{l}) + \sum_{v_j \in \mathcal{S}_{1} } (1-\bar{s})(x_{jl}-\bar{x}_{l})\right), \text{ where } \bar{A}:= \frac{1}{N}\sum_{v_i \in \mathcal{V} } a_{i} \text{ for any } a,\\
    &= \frac{1}{\bar{s} \sqrt{N \sigma^2}} \left(\sum_{v_i \in \mathcal{S}_{0} } \frac{-|\mathcal{S}_{1}|}{N}(x_{il}-\bar{x}_{l}) + \sum_{v_j \in \mathcal{S}_{1} }  \frac{|\mathcal{S}_{0}|}{N}(x_{jl}-\bar{x}_{l})\right),\\
    &= \frac{1}{\bar{s} \sqrt{N \sigma^2}} \left(\sum_{v_i \in \mathcal{S}_{0} } \frac{-|\mathcal{S}_{1}|}{N}x_{il} +  \frac{|\mathcal{S}_{1}|}{N} |\mathcal{S}_{0}|\bar{x}_{l}  + \sum_{v_j \in \mathcal{S}_{1} } \frac{|\mathcal{S}_{0}|}{N}x_{jl} -  \frac{|\mathcal{S}_{0}|}{N} |\mathcal{S}_{1}|\bar{x}_{l} \right),\\
    &=  \frac{1}{\bar{s} \sqrt{N \sigma^2}} \left(\sum_{v_i \in \mathcal{S}_{0} } \frac{-|\mathcal{S}_{1}|}{N}x_{il} +   \sum_{v_j \in \mathcal{S}_{1} } \frac{|\mathcal{S}_{0}|}{N}x_{jl}\right), \\
    &= \frac{|\mathcal{S}_{0}||\mathcal{S}_{1}|}{N^{3/2} \bar{s}\sigma} \left(\frac{1}{|\mathcal{S}_{1}|} \sum_{v_j \in \mathcal{S}_{1} }x_{jl} - \frac{1}{|\mathcal{S}_{0}|} \sum_{v_i \in \mathcal{S}_{0} }x_{il}\right).
    \end{split}
\end{equation}
Therefore , $\left\| \frac{1}{|\mathcal{S}_{0}|} \sum_{v_i \in \mathcal{S}_{0} } \mathbf{x}_{i} - \frac{1}{|\mathcal{S}_{0}|} \sum_{v_j \in \mathcal{S}_{1} } \mathbf{x}_{j}  \right\| = \frac{N^{3/2} \bar{s} \sigma}{|\mathcal{S}_{0}||\mathcal{S}_{1}|} \|\mathbf{\rho}\|$, implying
\begin{equation}
    \Delta^{raw}_{SP} \leq \frac{c N^{3/2} \bar{s} \sigma}{|\mathcal{S}_{0}||\mathcal{S}_{1}|\lambda} \| \boldsymbol{\rho} \|,
\end{equation}
which is the desired result. \hfil\qed

\section{Statement and Proof of Theorem \ref{theorem:nodal_feat_gpr}}
\label{app:proof_nodal_gpr}

\begin{theorem}
\label{theorem:nodal_feat_gpr} 
Consider nodal representations obtained via GPR. For the proposed node feature unlearning scheme under Assumptions i)-iv), 
it holds with high probability that:
\begin{equation}
    \|\nabla L(\tilde{\mathbf{w}}; \tilde{\mathcal{D}}_{G})\| \leqslant \frac{\gamma_{2} }{m}\left[ \frac{2c \sqrt{F} + c_{1}\sqrt{(F-k) m}}{\lambda\sqrt{F}}\right]^{2}.
\end{equation}
\end{theorem}

\textbf{Proof of Theorem \ref{theorem:nodal_feat_gpr}.} The argument follows the same general strategy as the proof we present in Appendix \ref{app:proof_nodal}. Thus, herein, we will only focus on the terms related to aggregated representations $\mathbf{z}_{i}$.

First, we show that $\|\mathbf{z}_{i}\| \leq 1$ holds with high probability. With GPR-based models, aggregated representations are obtained via $\|\mathbf{z}_{i}\| := \|\frac{1}{L+1} [\mathbf{x}_{i}, \mathbf{P}^{0}\mathbf{x}_{i}, \cdots, \mathbf{P}^{L}\mathbf{x}_{i}]\|$. Thus, it follows that
\begin{equation}
    \begin{split}
        \|\mathbf{z}_{i}\| &= \left\|\frac{1}{L+1} [\mathbf{x}_{i}, \mathbf{P}^{0}\mathbf{x}_{i}, \cdots, \mathbf{P}^{L}\mathbf{x}_{i}]\right\|,\\
       &= \frac{1}{L+1}\left\|[\mathbf{x}_{i}, \mathbf{P}^{0}\mathbf{x}_{i}, \cdots, \mathbf{P}^{L}\mathbf{x}_{i}]\right\|,\\
        & \leq \frac{1}{L+1} \sum_{l=0}^{L} \|\mathbf{P}^{l}\mathbf{x}_{i}\|.
    \end{split}
\end{equation}
\cite{chien2022certified} proves that $\|\mathbf{P}^{l}\mathbf{x}_{i}\| \leq 1$ for any $l=0, \cdots, L$ if $\|\mathbf{x}_{i}\| \leq 1$. Thus, with high probability, the following inequality holds,
\begin{equation}
    \begin{split}
        \|\mathbf{z}_{i}\|  & \leq \frac{1}{L+1} \sum_{l=0}^{L} \|\mathbf{P}^{l}\mathbf{x}_{i}\|,\\
       &\leq \frac{1}{L+1} L+1,\\
        &=1.
    \end{split}
\end{equation}
We also show in Appendix \ref{app:proof_nodal} that given $\|\mathbf{z}_{i}\| \leq 1$, it is guaranteed that $\|\tilde{\mathbf{z}}_{i}\| \leq 1$ for the proposed feature unlearning strategy. 

The second term that changes with GPR-based aggregation is $\left\|\sum_{v_i \in \mathcal{V}_{tr} } \ell^{\prime}\left(\tilde{\mathbf{z}}_i^{\top} \mathbf{w}^{*}, \mathbf{y}_i\right) (\mathbf{z}_i)_{n s}\right\|$, where we need to bound this term. To that end, we have
\begin{equation}
\begin{split}
     & \left\|\sum_{v_i \in \mathcal{V}_{tr} } \ell^{\prime}\left(\tilde{\mathbf{z}}_i^{\top} \mathbf{w}^{*}, \mathbf{y}_i\right) (\mathbf{z}_i)_{n s}\right\| \\
      &=  \left\|\sum_{v_i \in \mathcal{V}_{tr} } \frac{1}{L+1} \left[\sum_{v_j \in \mathcal{N}_{0}(v_{i})} \frac{\ell^{\prime}\left(\tilde{\mathbf{z}}_i^{\top} \mathbf{w}^{*}, \mathbf{y}_i\right) [\bar{A}^{0}]_{ij}}{d_{i}^{0}} (\mathbf{x}_j)_{n s}, \cdots, \sum_{v_j \in \mathcal{N}_{L}(v_{i})} \frac{\ell^{\prime}\left(\tilde{\mathbf{z}}_i^{\top} \mathbf{w}^{*}, \mathbf{y}_i\right) [\bar{A}^{L}]_{ij}}{d_{i}^{L}} (\mathbf{x}_j)_{n s} \right] \right\|.
\end{split}
\end{equation}
Let $\mathbf{o}:= \sum_{v_i \in \mathcal{V}_{tr} } \frac{1}{L+1} \left[\sum_{v_j \in \mathcal{N}_{0}(v_{i})} \frac{\ell^{\prime}\left(\tilde{\mathbf{z}}_i^{\top} \mathbf{w}^{*}, \mathbf{y}_i\right) [\bar{A}^{0}]_{ij}}{d_{i}^{0}} (\mathbf{x}_j)_{n s}, \cdots, \sum_{v_j \in \mathcal{N}_{L}(v_{i})} \frac{\ell^{\prime}\left(\tilde{\mathbf{z}}_i^{\top} \mathbf{w}^{*}, \mathbf{y}_i\right) [\bar{A}^{L}]_{ij}}{d_{i}^{L}} (\mathbf{x}_j)_{n s} \right]$. Similar to our proof in Appendix \ref{app:proof_nodal}, we can show that each element of $\mathbf{o}$ is a Gaussian RV, $o_{c} \sim \textrm{Normal}(0,\frac{1}{(L+1)^{2}} \sum_{v_i \in \mathcal{V}_{tr} } \sum_{v_j \in \mathcal{N}_{l}(v_{i})} \left[ \frac{\ell^{\prime}\left(\tilde{\mathbf{z}}_i^{\top} \mathbf{w}^{*}, \mathbf{y}_i\right) [\bar{A}^{l}]_{ij}}{deg_{i}^{l}}\right]^{2} \sigma^{2}), \forall c={1, \dots, (F-k)(L+1)}$. Therefore, the following inequalities hold with a probability of $0.9999994$:
\begin{equation}
\label{eq:ineq}
    \frac{-5}{L+1} \sqrt{\sum_{v_i \in \mathcal{V}_{tr} } \sum_{v_j \in \mathcal{N}_{l}(v_{i})} \left[ \frac{\ell^{\prime}\left(\tilde{\mathbf{z}}_i^{\top} \mathbf{w}^{*}, \mathbf{y}_i\right) [\bar{A}^{l}]_{ij}}{d_{i}^{l}}\right]^{2}\sigma^{2}} \leq o_{l} \leq \frac{5}{L+1} \sqrt{\sum_{v_i \in \mathcal{V}_{tr} } \sum_{v_j \in \mathcal{N}_{l}(v_{i})} \left[ \frac{\ell^{\prime}\left(\tilde{\mathbf{z}}_i^{\top} \mathbf{w}^{*}, \mathbf{y}_i\right) [\bar{A}^{l}]_{ij}}{d_{i}^{l}}\right]^{2}\sigma^{2}}.
\end{equation}
5
Thus, we can derive the same bound as in the proof presented in Appendix \ref{app:proof_nodal} by the following argument:
\begin{equation*}
\begin{split}
     \|\mathbf{o}\| &\leq \sqrt{\frac{25(F-k)(L+1)}{(L+1)^{2}} \sum_{v_i \in \mathcal{V}_{tr} } \sum_{v_j \in \mathcal{N}_{l}(v_{i})} \left[ \frac{\ell^{\prime}\left(\tilde{\mathbf{z}}_i^{\top} \mathbf{w}^{*}, \mathbf{y}_i\right) [\bar{A}^{l}]_{ij}}{d_{i}^{l}}\right]^{2} \sigma^{2} }, \text{ for any } l={0, \cdots, L},\\
     &\leq \sqrt{25(F-k) \sigma^{2} \sum_{v_i \in \mathcal{V}_{tr} }(\ell^{\prime}\left(\tilde{\mathbf{z}}_i^{\top} \mathbf{w}^{*}, \mathbf{y}_i\right))^{2}}, \\
     &\leq c_{1} \sqrt{\frac{(F-k)}{F} m}
\end{split}
\end{equation*}
and the result follows.\hfill\qed

\section{Additional Dataset Details}
\label{app:dataset}

As our work approaches the graph unlearning problem from a fairness perspective, we employ commonly used real-world data for fair graph ML tasks. Specifically, the results are reported on four attributed networks: Credit Defaulter \cite{agarwal2021towards}, Recidivism \cite{agarwal2021towards},  German Credit \cite{agarwal2021towards}, Pokec-z \cite{dai2021say}. Credit Defaulter consists of $N=30,000$ nodes, where the sensitive attribute is age and the classification target is to predict whether an individual will default on the credit card payment. The Recidivism network has $N=18,876$ nodes representing defendants who got released on bail at US state courts, and the goal is to predict bail decision with race as the sensitive attribute. German Credit has $N=1,000$ nodes that are clients in a bank, here the classification labels are good
vs. bad credit scores and gender is used as the sensitive attribute. Finally, Pokec-z is sampled from an anonymized version of the Pokec network in 2012 (a social network from Slovakia), where nodes correspond to users who live in two major regions, and the region information is utilized as the sensitive attribute \cite{dai2021say}.

Statistical information for these datasets is presented in Table \ref{table:stats}, where $F$ is the total number of nodal features. In Table \ref{table:stats}, $|\mathcal{E}^{\chi}|$ and $|\mathcal{E}^{\omega}|$ represent the number of inter-edges and intra-edges, respectively.

\begin{table}[h]
	\centering
\caption{Dataset statistics. }
\label{table:stats}
\begin{footnotesize}
\resizebox{0.9\textwidth}{!}{
\begin{tabular}{c c c c c c c c}
\toprule
                                                    Dataset &  $|\mathcal{V}|$ &
                           $|\mathcal{E}|$ & $F$ & $|\mathcal{S}_{0}|$ &$|\mathcal{S}_{1}|$ & $|\mathcal{E}^{\chi}|$&$|\mathcal{E}^{\omega}|$\\ 
\midrule
Credit Defaulter  & $30000$ & $304754$ & $13$ & $27315$ & $2685$ & $33254$ & $271500$\\
Recidivism & $18876$ & $642616$ & $18$ & $9317$ & $9559$ & $298098$ & $344518$ \\
German Credit & $1000$ & $44484$ & $27$ & $690$ & $310$ & $21492$ & $22992$ \\
Pokec-z & $10262$ & $51786$ & $277$ & $6617$ & $3645$ & $1772$ & $50014$\\
\bottomrule
\end{tabular}}
\end{footnotesize}
\end{table}

\section{Additional Results for Node Feature Unlearning}
\label{app:add_feat_res}
\begin{figure}[h!]
    \centering
    \subfigure[Credit Defaulter]{{\includegraphics[width=6.85cm]{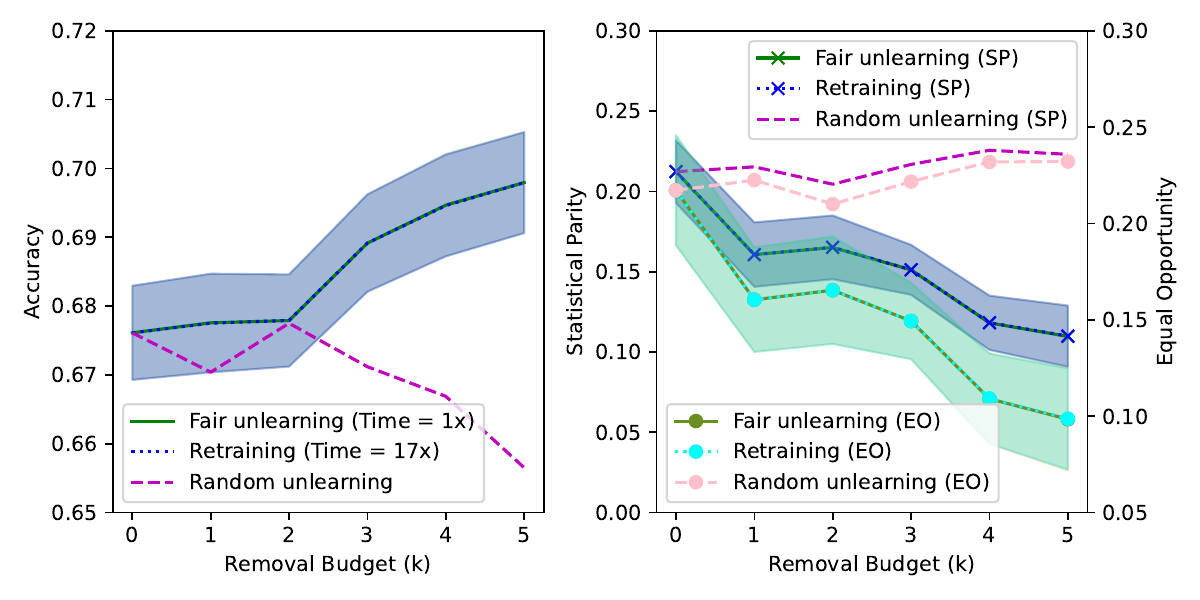} }}%
    \hspace{0.0cm}
    \subfigure[German Credit]{{\includegraphics[width=6.85cm]{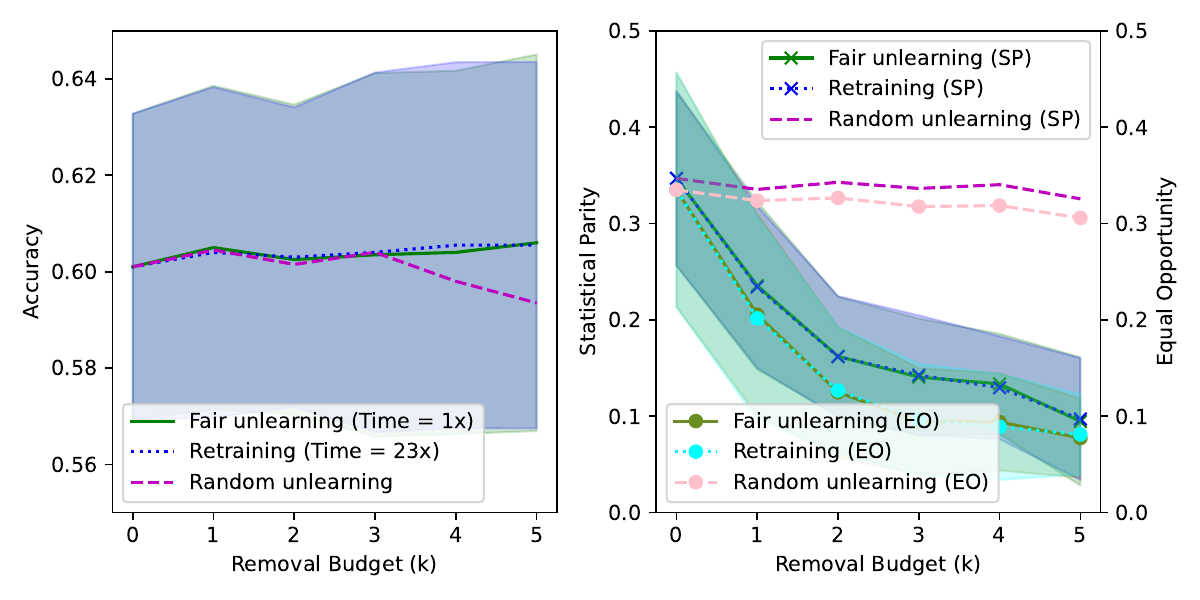} }}%
    \caption{Feature unlearning with fairness-agnostic (random) and proposed unlearning mechanisms.}%
    \label{fig:nodal_feat1}%
\end{figure}

\begin{figure}[h!]
    \centering
    \subfigure[Recidivism]{{\includegraphics[width=6.85cm]{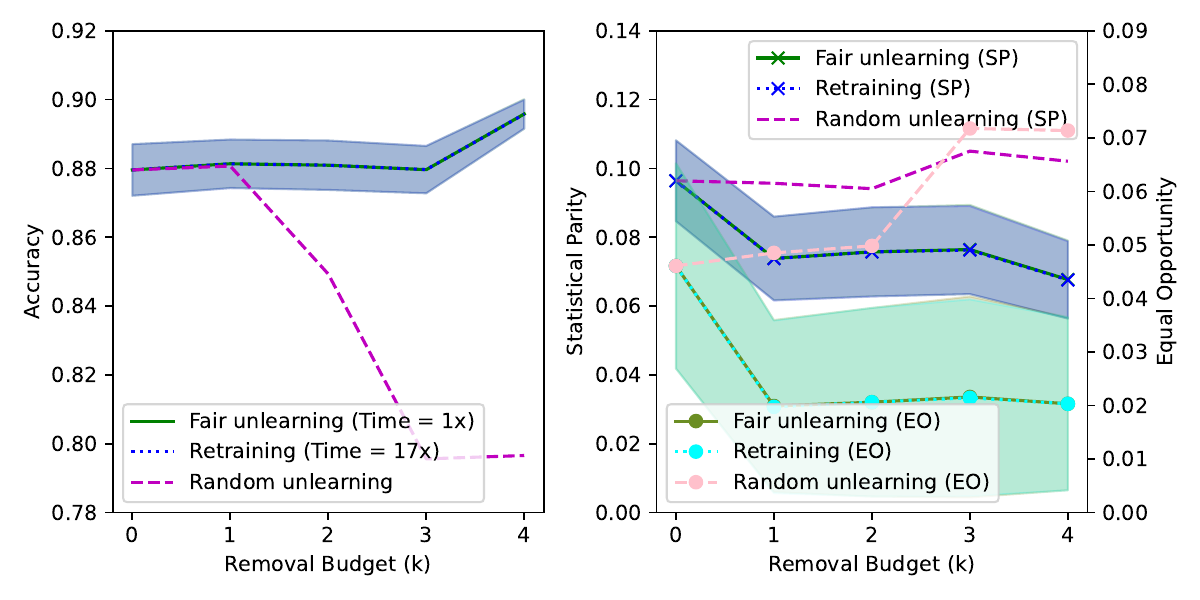} }}%
    \hspace{0.0cm}
    \subfigure[Pokec-z]{{\includegraphics[width=6.85cm]{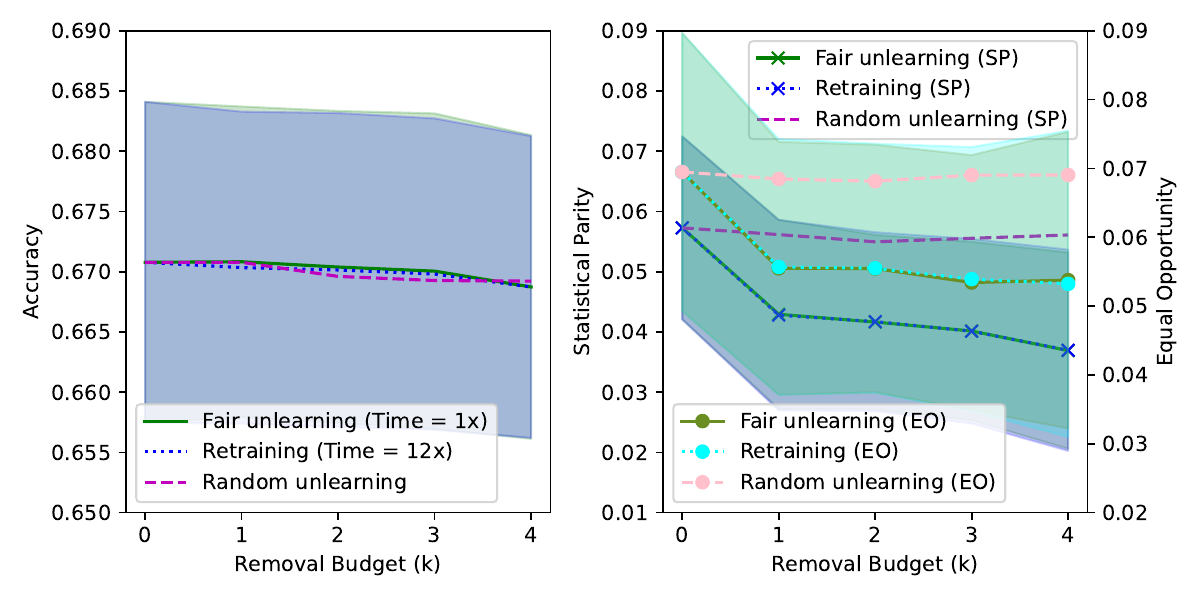} }}%
    \caption{Feature unlearning with fairness-agnostic (random) and proposed unlearning mechanisms.}%
    \label{fig:nodal_feat2}%
\end{figure}

Figures \ref{fig:nodal_feat1} and \ref{fig:nodal_feat2} report classification accuracy, fairness performance in terms of statistical parity and equal opportunity together with the runtime comparison between retraining from scratch and our proposed update. All in all, the reported performance in these figures suggest that our strategy can reduce bias metrics significantly
without sacrificing utility, and at a fraction of the computational complexity.

\section{Additional Results for Edge Unlearning}
\label{app:add_edge_res}
\begin{figure}[h]
    \centering
    \subfigure[Recidivism]{{\includegraphics[width=6.85cm]{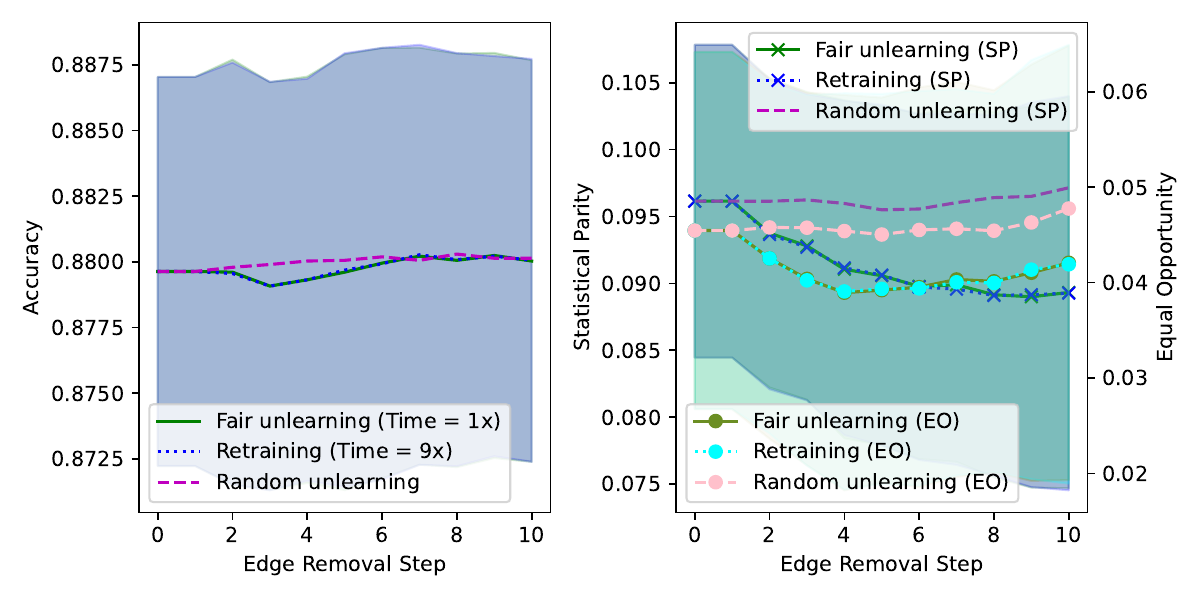} }}%
    \hspace{0.0cm}
    \subfigure[Pokec-z]{{\includegraphics[width=6.85cm]
    {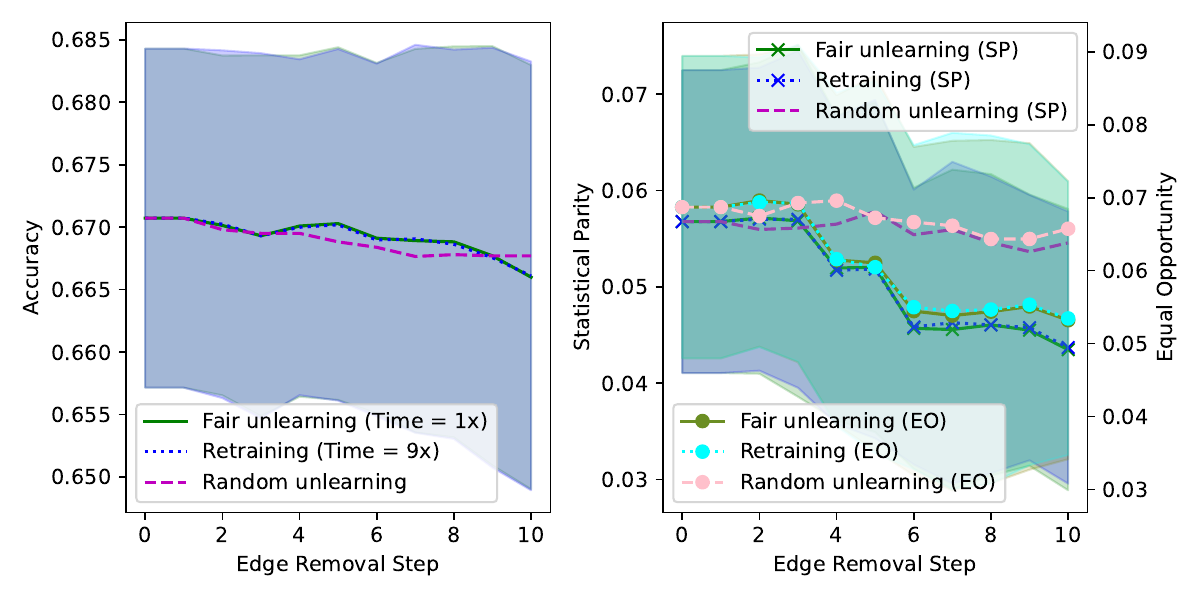} }}%
    \caption{Edge unlearning with fairness-agnostic (random) and proposed unlearning mechanisms.}%
    \label{fig:edge2}%
\end{figure}

Figure \ref{fig:edge2} presents comparative node classification results for the proposed edge unlearning strategy, random unlearning, and retraining from scratch. Similar to Fig. \ref{fig:edge1}, results in Fig. \ref{fig:edge2} for Recidivism and Pokec-z demonstrate the favorable fairness-utility trade-off with significantly lower runtimes compared to retraining from scratch.

\begin{figure}[h]
    \centering
    \subfigure[Credit Defaulter]{{\includegraphics[width=6.85cm]{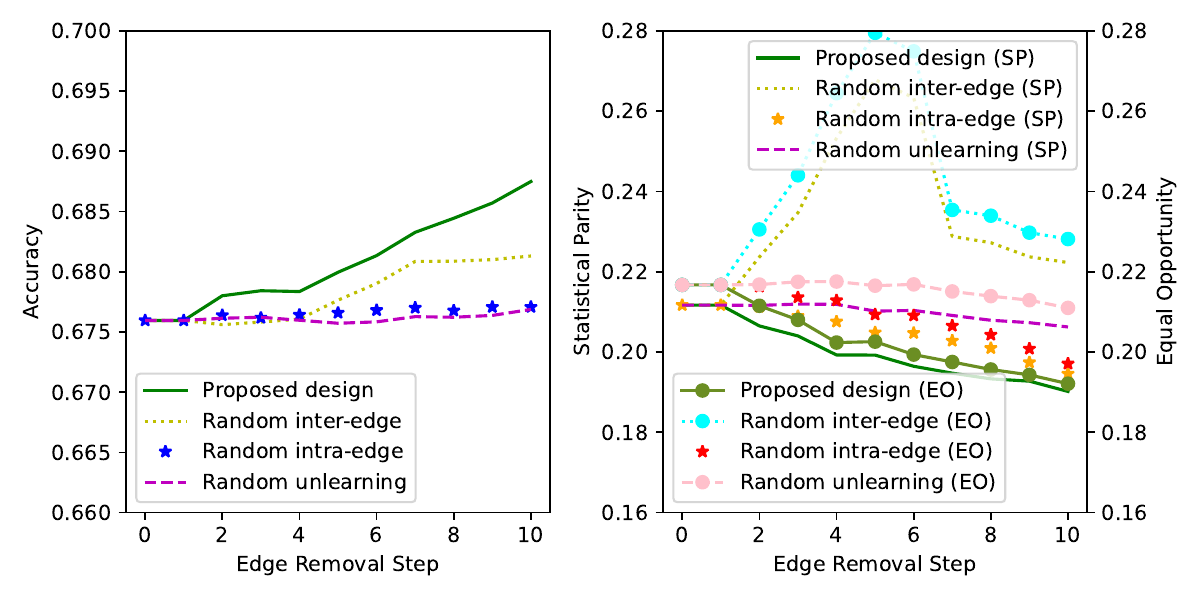} }}%
    \hspace{0.0cm}
    \subfigure[German Credit]{{\includegraphics[width=6.85cm]{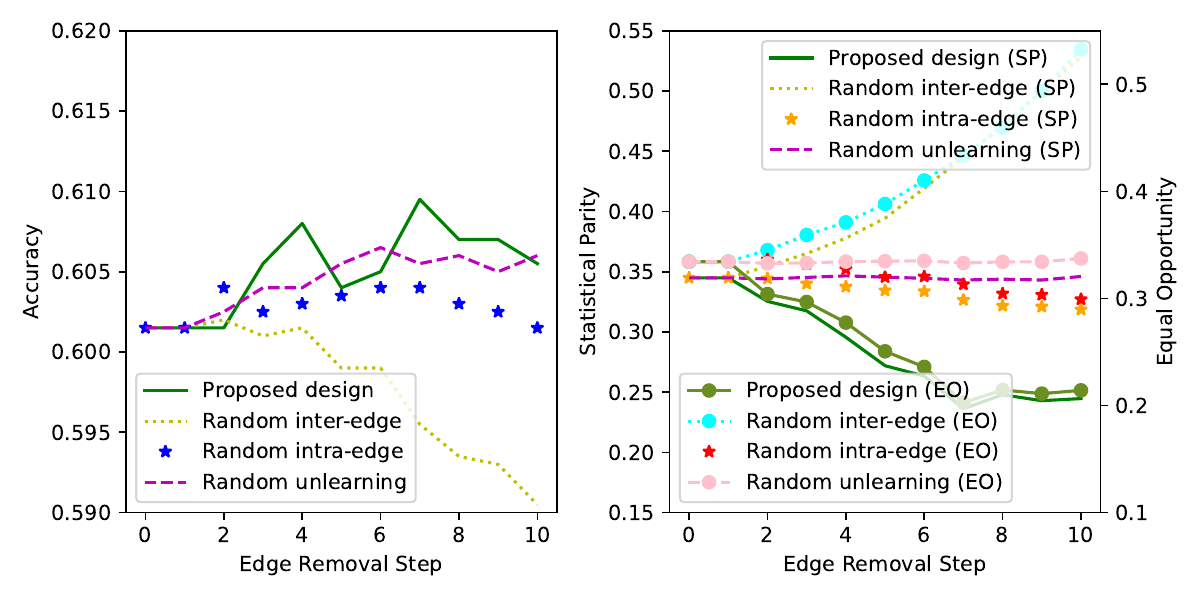} }}%
    \caption{Comparison of different edge selection mechanisms.}%
    \label{fig:edge_ablation1}%
\end{figure}

\begin{figure}[h]
    \centering
    \subfigure[Recidivism]{{\includegraphics[width=6.85cm]{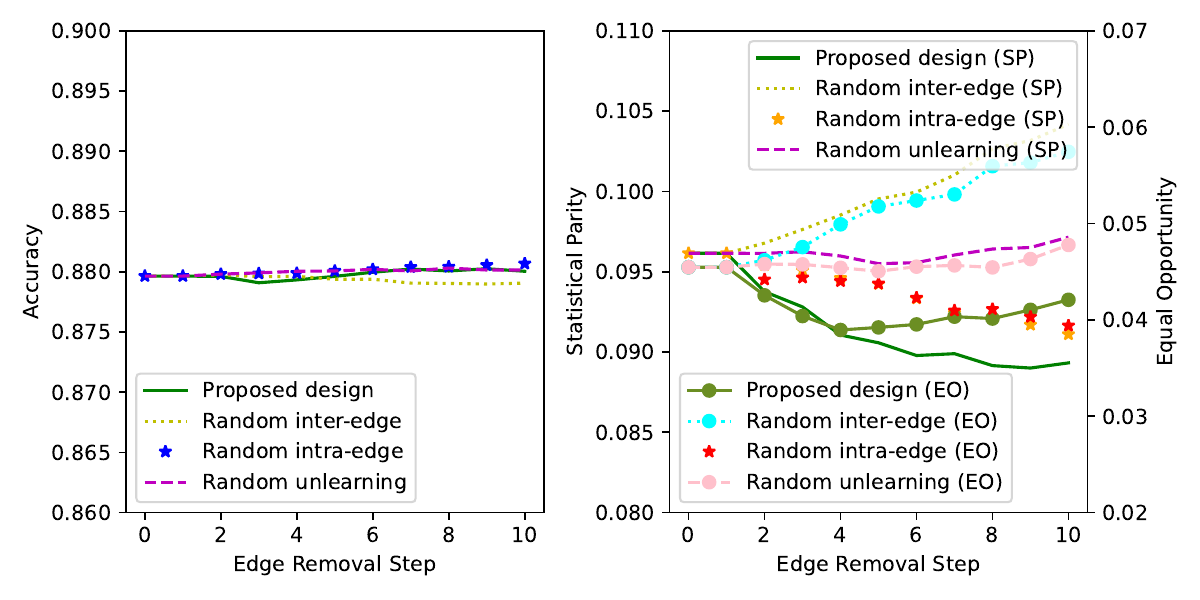} }}%
    \hspace{0.0cm}
    \subfigure[Pokec-z]{{\includegraphics[width=6.85cm]{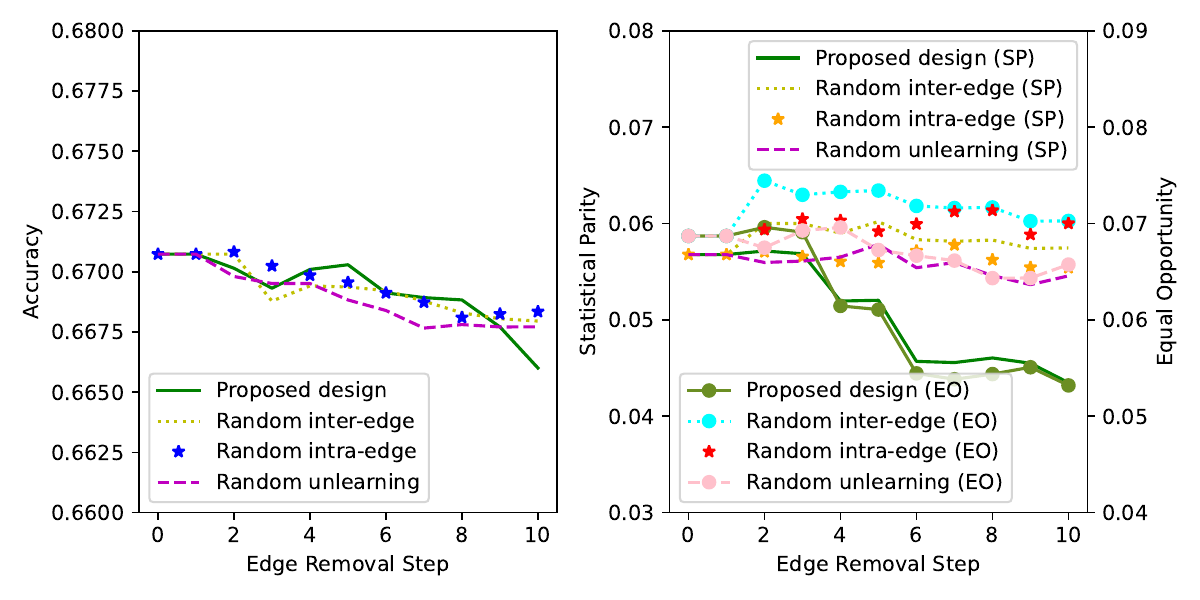} }}%
    \caption{Comparison of different edge selection mechanisms.}%
    \label{fig:edge_ablation2}%
\end{figure}

\textbf{Ablation Study.} In order to demonstrate the impacts of design choices in bias score $b_{e}$ presented in Section \ref{subsec:struct}, in addition to the proposed strategy and random feature selection, we report the utility/fairness performances for; (i) random selection of intra-edges; and (ii) random selection of inter-edges. Overall, Figures \ref{fig:edge_ablation1} and \ref{fig:edge_ablation2} show that unlearning inter-edges deteriorates the fairness metrics, supporting the findings in \eqref{eq:tnnls}, and our design choice of focusing on intra-edges. Furthermore, prioritizing the edges incident to low-degree nodes can indeed mitigate bias more effectively compared to random intra-edge unlearning, supporting our design choice of making $b_e$ a function of $\operatorname{min}(d_{i}, d_{j})$.

\section{Additional Results for Node Unlearning}
\label{app:add_node_res}

\begin{figure}[h]
    \centering
    \subfigure[Credit Defaulter]{{\includegraphics[width=6.85cm]{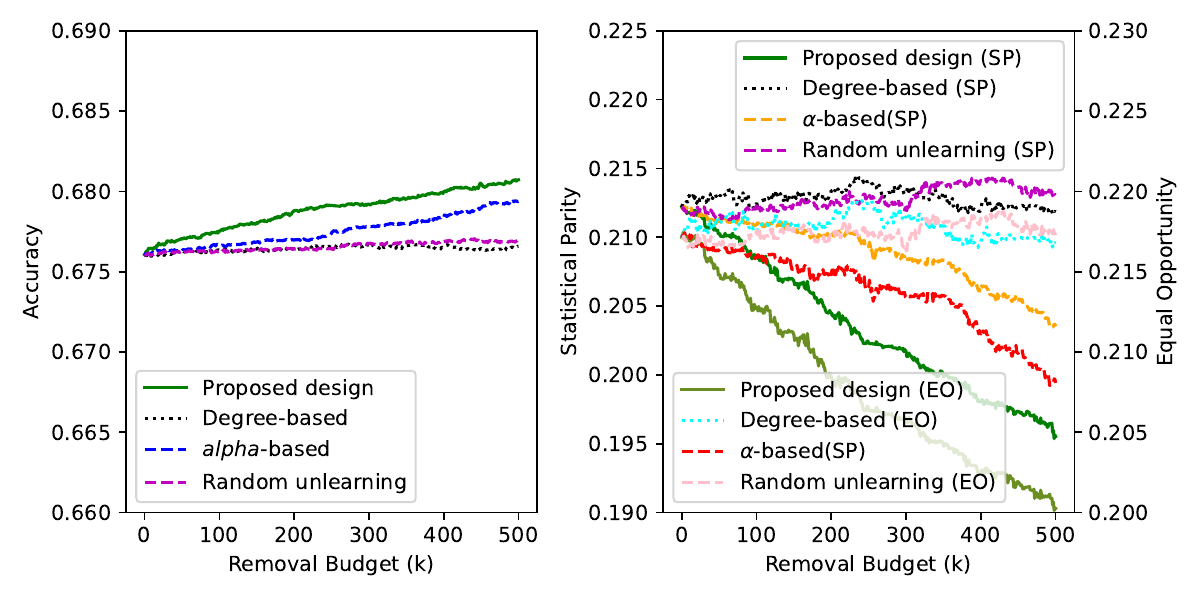} }}%
    \hspace{0.0cm}
    \subfigure[German Credit]{{\includegraphics[width=6.85cm]{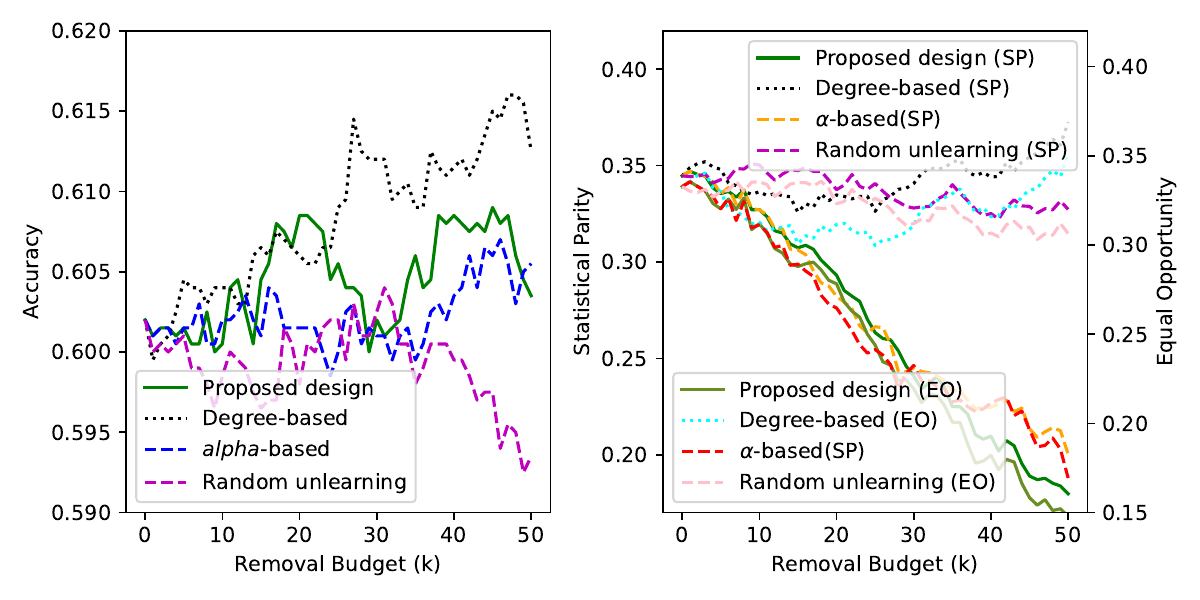} }}%
    \caption{Comparison of different node selection mechanisms.}%
    \label{fig:node_ablation1}%
\end{figure}

For node unlearning, the proposed bias score used in node selection, $b_{n}$, is composed of two components, (i) a bias related component $\frac{d_{i}^{\omega}}{1 + d_{i}^{\chi}}$; (ii) a utility-centered design element $\frac{1}{d_{i}}$. Herein, we design an ablation study to understand the individual effects of these terms. Specifically, in Figure \ref{fig:node_ablation1}, we report the results for the proposed node selection strategy, random node selection, node selection with $\frac{d_{i}^{\omega}}{1 + d_{i}^{\chi}}$ as the bias score that is inspired by $\alpha$-related bias terms in \eqref{eq:tnnls}, and node selection by using degree-based component $\frac{1}{d_{i}}$ as the bias score. All in all, the results show that considering bias factor $\frac{d_{i}^{\omega}}{1 + d_{i}^{\chi}}$ together with utility-oriented component $\frac{1}{d_{i}}$ provides the best fairness/utility trade-off. 


\section{Experimental Setup and Hyperparameters}
\label{app:hypers}
As the optimizer, we employ L-BFGS \cite{liu1989limited} based on its reported efficiency on strongly convex problems~\cite{guo2020certified}. Unless specified
otherwise, in the experiments, we fix $\delta = 1e-4, \epsilon = 1, \lambda = 10$ for all experiments (generally following \cite{chien2022certified}), and $L$ is selected on the basis of node classification accuracy, where different hop numbers $L\in\{2, 3, 4, 5, 6\}$ are considered. For training, we use $60\%$ of the nodes, while the remaining data are split equally among the validation and test sets. Results are obtained over 10 different data splits and their averages are reported along with the standard deviations. 

\section{Computational Resources}
\label{app:resources}
Experiments are conducted using 4 NVIDIA RTX A4000 GPUs.

\end{document}